\documentclass{article}

\usepackage{microtype}
\usepackage{graphicx}
\usepackage{booktabs}
\usepackage{array}
\newcolumntype{L}[1]{>{\raggedright\arraybackslash}p{#1}}
\usepackage{hyperref}

\usepackage[preprint]{icml2026}
\hypersetup{pdfsubject={Preprint}}

\usepackage{amsmath}
\usepackage{amssymb}
\usepackage{mathtools}
\usepackage{amsthm}
\usepackage{xcolor}
\usepackage{tikz}
\usetikzlibrary{arrows.meta,positioning,fit,calc}
\usepackage{pgfplots}
\pgfplotsset{compat=1.16}
\usepackage{fontawesome5}
\usepackage{colortbl}
\usepackage{listings}
\usepackage{tcolorbox}
\tcbuselibrary{skins,listings}
\usepackage{flafter}
\usepackage[capitalize,noabbrev]{cleveref}

\renewcommand{\algorithmicrequire}{\textbf{Input:}}
\renewcommand{\algorithmicensure}{\textbf{Output:}}
\newcommand{\acmt}[1]{\hfill{$\triangleright$~#1}}

\newcommand{\grp}[1]{\rowcolor{kviolet!9}\multicolumn{4}{@{}l@{}}{\rule{0pt}{2.4ex}\textcolor{kviolet!80!black}{\textbf{#1}}}\\}
\newcommand{\loc}[2]{\texttt{L#1\,n#2}}
\newcommand{\ba}[2]{$#1\!\to\!#2$}

\lstdefinestyle{vlog}{language=Verilog, basicstyle=\ttfamily\fontsize{6.6}{7.9}\selectfont, columns=fullflexible,
  keepspaces=true, keywordstyle=\color{kviolet!85!black}\bfseries, commentstyle=\color{kmuted}\itshape,
  numberstyle=\color{kmuted}\fontsize{5.5}{7}\selectfont, showstringspaces=false, upquote=true, breaklines=true, breakindent=12pt, aboveskip=1pt, belowskip=1pt}
\newtcolorbox{chatbox}[2]{enhanced, colback=#1!5, colframe=#1!70!black, boxrule=0.5pt, arc=2pt,
  left=3pt, right=3pt, top=2.4mm, bottom=1.2pt, fontupper=\scriptsize, before skip=3.5mm, after skip=0.6mm,
  title={#2}, fonttitle=\scriptsize\bfseries, coltitle=white, colbacktitle=#1!70!black,
  attach boxed title to top left={xshift=2mm, yshift=-2mm},
  boxed title style={arc=1pt, boxrule=0pt, top=0.2pt, bottom=0.2pt, left=2pt, right=2pt}}

\theoremstyle{plain}
\newtheorem{proposition}{Proposition}[section]
\crefname{proposition}{Proposition}{Propositions}
\crefname{appendix}{Appendix}{Appendices}
\Crefname{appendix}{Appendix}{Appendices}
\crefname{figure}{Fig.}{Figs.}
\Crefname{figure}{Figure}{Figures}

\newcommand{\E}{\mathbb{E}}
\newcommand{\Prob}{\mathbb{P}}
\newcommand{\T}{^{\top}}
\newcommand{\silu}{\operatorname{silu}}
\newcommand{\sys}{Kolpona}
\newcommand{\sysmini}{Kolpona-mini}

\definecolor{kblue}{HTML}{2A78D6}
\definecolor{korange}{HTML}{EB6834}
\definecolor{kaqua}{HTML}{1BAF7A}
\definecolor{kyellow}{HTML}{EDA100}
\definecolor{kviolet}{HTML}{6250D6}
\definecolor{kred}{HTML}{E34948}
\definecolor{kink}{HTML}{0B0B0B}
\definecolor{kinkb}{HTML}{3A3936}
\definecolor{kmuted}{HTML}{5A5954}
\definecolor{kgrid}{HTML}{E1E0D9}
\definecolor{kaxis}{HTML}{C3C2B7}
\definecolor{kpub}{HTML}{BDBBB2}
\newcommand{\figfont}{\sffamily\fontsize{7}{8.4}\selectfont}
\newcommand{\figfontS}{\sffamily\fontsize{6.2}{7.4}\selectfont}
\newcommand{\figfontT}{\sffamily\bfseries\fontsize{7.6}{9}\selectfont}
\newcommand{\figtt}[1]{{\fontfamily{txtt}\fontsize{6.4}{7.4}\selectfont #1}}
\tikzset{
  kindA/.style={fill=kblue},
  kindB/.style={fill=korange},
  kindP/.style={fill=kpub},
  kinda/.style={fill=white, draw=kblue, line width=0.7pt},
  kindb/.style={fill=white, draw=korange, line width=0.7pt},
  nameA/.style={text=kink, font=\figfontS\bfseries},
  nameB/.style={text=kink, font=\figfontS\bfseries},
  nameP/.style={text=kinkb, font=\figfontS},
  namea/.style={text=kinkb, font=\figfontS},
  nameb/.style={text=kinkb, font=\figfontS},
  valA/.style={text=kink, font=\figfontS\bfseries},
  valB/.style={text=kink, font=\figfontS\bfseries},
  valP/.style={text=kinkb, font=\figfontS},
  vala/.style={text=kinkb, font=\figfontS},
  valb/.style={text=kinkb, font=\figfontS},
}

\icmltitlerunning{\sys{}: An Agentic Framework with Gradient-Free Neuron Programming for RTL Generation}

\newsavebox{\exbox}   %
\begin{document}
\raggedbottom
\clubpenalty=10000 \widowpenalty=10000 \displaywidowpenalty=10000 \brokenpenalty=10000

\twocolumn[
  \icmltitle{\sys{}: An Agentic Framework with Gradient-Free\\ Neuron Programming for RTL Generation}

  \begin{icmlauthorlist}
    \icmlauthor{Musa Tur Farazi}{cse}
    \icmlauthor{Md Nasir Uddin}{cse}
    \icmlauthor{Soumik Bhattacharjee}{cse}
    \icmlauthor{Tishna Sabrina}{uap}
    \icmlauthor{Chowdhury Mohammad Rakin Haider}{cse}
  \end{icmlauthorlist}

  \icmlaffiliation{cse}{Department of Computer Science and Engineering, Bangladesh University of Engineering and Technology (BUET), Dhaka-1000, Bangladesh}
  \icmlaffiliation{uap}{Department of Electrical and Electronic Engineering, University of Asia Pacific, Dhaka, Bangladesh}
  \icmlcorrespondingauthor{Musa Tur Farazi}{2005038@cse.buet.ac.bd, musatur330@gmail.com}

  \icmlkeywords{RTL generation, Verilog, neuron programming, weight-space programming, model editing, test-time selection, agents}

  \vskip 0.3in
]

\printAffiliationsAndNotice{}

\begin{abstract}
Open-weight large language models (LLMs) for register-transfer level (RTL) design are usually improved by fine-tuning on curated Verilog corpora, which needs large sets of verified designs and GPU training. We present \sys{}, an agentic framework that improves such models without gradient training. Its first component is neuron programming, a new form of weight-space programming (WSP) that writes single neurons in closed form into the weights to remove failure habits found on practice tasks, and keeps each edit only when testbench results on fresh tasks favour it. At inference time, tool-using agents propose designs, and a label-free selection stage delivers one design per task from the specification, the candidates and their simulations alone. On Qwen3.5-4B, edits to 16 neurons raise the VerilogEval v2 pass@1 of the model alone from 31.7\% to 55.4\%, and the full framework reaches 77.6\%. On Qwen3.8-27B, \sys{} lifts pass@1 from 73.1\% to 94.9\% on VerilogEval v2 and from 66.0\% to 92.0\% on RTLLM v2.1 on a single 32\,GB GPU, above every published system in our comparison. The RTL edits also carry over to Python and raise the pass@1 of the 4B model alone from 84.8\% to 91.5\% on HumanEval+ and from 67.7\% to 77.5\% on MBPP+.
\end{abstract}

\section{Introduction}
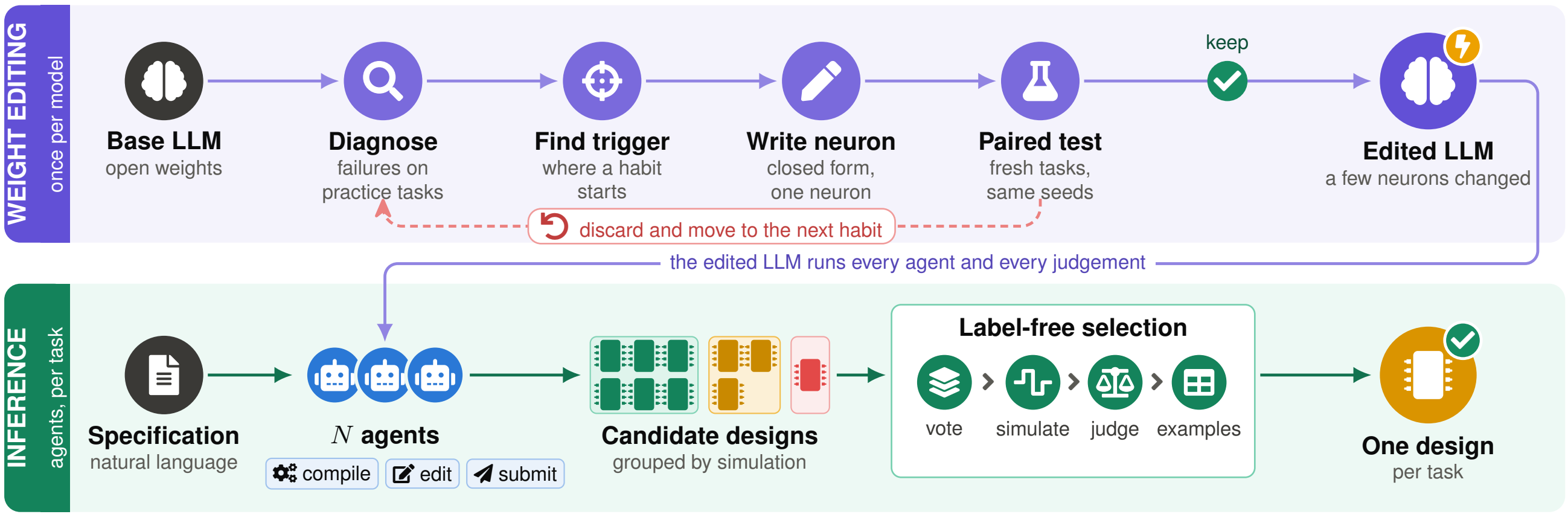
\begin{figure*}[t]
\centering
\begin{tikzpicture}[x=1cm,y=1cm,
    lbl/.style={font=\figfontT, text=kink, anchor=north, inner sep=0pt, align=center},
    sub/.style={font=\figfontS, text=kmuted, anchor=north, inner sep=0pt, align=center},
    flow/.style={-{Latex[length=2mm,width=1.8mm]}, line width=0.9pt, draw=#1},
    tool/.style={font=\figfontS, text=kink, fill=kblue!10, draw=kblue!45, line width=0.4pt, rounded corners=1.8pt,
                 inner xsep=2.2pt, inner ysep=1.5pt, anchor=center}]
  \def\icon{\fontsize{12.5}{12.5}\selectfont}
  \def\iconS{\fontsize{8.6}{8.6}\selectfont}
  \newcommand{\kb}[6]{%
    \node[circle, minimum size=0.86cm, inner sep=0pt, fill=#3, text=white, font=\icon] at (#1,#2) {#4};
    \node[lbl] at (#1,#2-0.55) {#5};
    \node[sub] at (#1,#2-0.86) {#6};}
  \path[fill=kviolet!7, rounded corners=5pt] (0,2.95) rectangle (17.1,5.55);
  \path[fill=kaqua!8, rounded corners=5pt] (0,0) rectangle (17.1,2.5);
  \path[fill=kviolet, rounded corners=5pt] (0,2.95) rectangle (0.72,5.55);
  \path[fill=kviolet] (0.4,2.95) rectangle (0.72,5.55);
  \path[fill=kaqua!75!black, rounded corners=5pt] (0,0) rectangle (0.72,2.5);
  \path[fill=kaqua!75!black] (0.4,0) rectangle (0.72,2.5);
  \node[rotate=90, text=white, align=center, inner sep=0pt] at (0.36,4.25)
    {{\figfontT WEIGHT EDITING}\\[-0.5pt]{\figfontS once per model}};
  \node[rotate=90, text=white, align=center, inner sep=0pt] at (0.36,1.25)
    {{\figfontT INFERENCE}\\[-0.5pt]{\figfontS agents, per task}};
  \kb{1.75}{4.72}{kinkb}{\faBrain}{Base LLM}{open weights}
  \kb{4.15}{4.72}{kviolet!85}{\faSearch}{Diagnose}{failures on\\practice tasks}
  \kb{6.55}{4.72}{kviolet!85}{\faCrosshairs}{Find trigger}{where a habit\\starts}
  \kb{8.95}{4.72}{kviolet!85}{\faPen}{Write neuron}{closed form,\\one neuron}
  \kb{11.35}{4.72}{kviolet!85}{\faFlask}{Paired test}{fresh tasks,\\same seeds}
  \foreach \xa/\xb in {2.23/3.67, 4.63/6.07, 7.03/8.47, 9.43/10.87} { \draw[flow=kviolet!70] (\xa,4.72) -- (\xb,4.72); }
  \draw[flow=kviolet!70] (11.83,4.72) -- (14.98,4.72);
  \node[circle, minimum size=0.44cm, inner sep=0pt, fill=kaqua!80!black, text=white, font=\iconS] at (13.4,4.72) {\faCheck};
  \node[font=\figfontS, text=kaqua!45!black, anchor=south, inner sep=2pt] at (13.4,4.95) {keep};
  \node[circle, minimum size=1.06cm, inner sep=0pt, fill=kviolet, text=white, font=\fontsize{15}{15}\selectfont] at (15.6,4.72) {\faBrain};
  \node[circle, minimum size=0.4cm, inner sep=0pt, fill=kyellow, draw=white, line width=0.9pt, text=white, font=\fontsize{7.5}{7.5}\selectfont] at (15.98,5.1) {\faBolt};
  \node[lbl] at (15.6,4.06) {Edited LLM};
  \node[sub] at (15.6,3.75) {a few neurons changed};
  \draw[kred!70, line width=0.9pt, dash pattern=on 2.4pt off 1.6pt, rounded corners=6pt,
        -{Stealth[length=2.6mm, width=2.2mm, round]}] (11.35,3.43) -- (11.35,3.13) -- (4.15,3.13) -- (4.15,3.43);
  \node[fill=white, draw=kred!55, line width=0.5pt, rounded corners=4pt, inner xsep=4pt, inner ysep=1.2pt,
        font=\figfontS, text=kred!85!black] at (7.75,3.13)
    {{\iconS\faUndo}\hspace{1.2mm}discard and move to the next habit};
  \kb{1.75}{1.5}{kinkb}{\faFile*}{Specification}{natural language}
  \foreach \x in {3.62,4.17,4.72} {
    \node[circle, minimum size=0.64cm, inner sep=0pt, fill=kblue, draw=kaqua!8, line width=1.1pt, text=white,
          font=\fontsize{10}{10}\selectfont] at (\x,1.5) {\faRobot};
  }
  \node[lbl] at (4.17,0.95) {$N$ agents};
  \def\iconT{\fontsize{6}{6}\selectfont}
  \node[tool, anchor=west] (t1) at (2.86,0.42) {{\iconT\faCogs}\hspace{0.6mm}compile};
  \node[tool, anchor=west] (t2) at ($(t1.east)+(0.07,0)$) {{\iconT\faEdit}\hspace{0.6mm}edit};
  \node[tool, anchor=west] (t3) at ($(t2.east)+(0.07,0)$) {{\iconT\faPaperPlane}\hspace{0.6mm}submit};
  \path[fill=kaqua!14, draw=kaqua!55, line width=0.4pt, rounded corners=2pt] (6.42,1.08) rectangle (7.6,1.92);
  \path[fill=kyellow!16, draw=kyellow!70, line width=0.4pt, rounded corners=2pt] (7.72,1.08) rectangle (8.5,1.92);
  \path[fill=kred!10, draw=kred!50, line width=0.4pt, rounded corners=2pt] (8.62,1.08) rectangle (9.04,1.92);
  \foreach \x in {6.64,7.01,7.38} { \foreach \y in {1.71,1.29} {
    \node[text=kaqua!75!black, font=\fontsize{10}{10}\selectfont, inner sep=0pt] at (\x,\y) {\faMicrochip}; } }
  \foreach \x/\y in {7.93/1.71, 8.29/1.71, 7.93/1.29} {
    \node[text=kyellow!85!black, font=\fontsize{10}{10}\selectfont, inner sep=0pt] at (\x,\y) {\faMicrochip}; }
  \node[text=kred, font=\fontsize{10}{10}\selectfont, inner sep=0pt] at (8.83,1.5) {\faMicrochip};
  \node[lbl] at (7.73,0.95) {Candidate designs};
  \node[sub] at (7.73,0.64) {grouped by simulation};
  \path[fill=white, draw=kaqua!60, line width=0.5pt, rounded corners=3pt] (9.72,0.38) rectangle (13.7,2.27);
  \node[font=\figfontT, text=kink, anchor=north, inner sep=0pt] at (11.71,2.13) {Label-free selection};
  \foreach \x/\ic/\t in {10.3/\faLayerGroup/vote, 11.27/\faWaveSquare/simulate, 12.17/\faBalanceScale/judge, 13.09/\faTable/examples} {
    \node[circle, minimum size=0.6cm, inner sep=0pt, fill=kaqua!75!black, text=white, font=\fontsize{9.5}{9.5}\selectfont] at (\x,1.42) {\ic};
    \node[font=\figfontS, text=kinkb, anchor=north, inner sep=0pt] at (\x,1.0) {\t};
  }
  \foreach \x in {10.78, 11.72, 12.63} { \node[text=kmuted, font=\iconS, inner sep=0pt] at (\x,1.42) {\faAngleRight}; }
  \node[circle, minimum size=1.06cm, inner sep=0pt, fill=kyellow!92!black, text=white, font=\fontsize{15}{15}\selectfont] at (15.6,1.5) {\faMicrochip};
  \node[circle, minimum size=0.4cm, inner sep=0pt, fill=kaqua!80!black, draw=white, line width=0.9pt, text=white, font=\fontsize{7.5}{7.5}\selectfont] at (15.98,1.88) {\faCheck};
  \node[lbl] at (15.6,0.84) {One design};
  \node[sub] at (15.6,0.53) {per task};
  \foreach \xa/\xb in {2.23/3.16, 5.1/6.32, 9.12/9.66, 13.76/14.98} { \draw[flow=kaqua!65!black] (\xa,1.5) -- (\xb,1.5); }
  \draw[flow=kviolet!70, rounded corners=4pt] (16.16,4.72) -- (16.8,4.72) -- (16.8,2.72) -- (4.17,2.72) -- (4.17,1.86);
  \node[fill=white, inner xsep=3pt, inner ysep=0.5pt, font=\figfontS, text=kviolet!85!black] at (9.9,2.72)
    {the edited LLM runs every agent and every judgement};
\end{tikzpicture}
\caption{Overview of \sys{}. Neuron programming writes single MLP neurons that remove failure habits found on practice tasks, and it keeps an edit only if a paired testbench comparison favours it. At inference time, the edited model runs $N$ tool-using agents and a label-free selection stage, which delivers one design per task without any testbench or reference design.}
\label{fig:overview}
\end{figure*}

Large language models (LLMs) can now write synthesizable register-transfer level (RTL) code from natural-language specifications \citep{liu2023verilogeval,lu2024rtllm,pinckney2025cvdp}. Most of the progress on open-weight models comes from fine-tuning on curated Verilog corpora with supervised or reinforcement learning \citep{liu2024rtlcoder,cui2024origen,liu2025craftrtl,zhu2025codevr1,yu2026chipmate}, often combined with agents that compile and simulate their designs \citep{zhao2025mage,ho2025verilogcoder}. Fine-tuning needs large sets of verified designs and GPU training, and it usually changes every weight of the model.

Weight-space programming (WSP) takes a different route. A capable LLM reads the activations and weights of a target model and edits its weights directly, using forward passes only \citep{takehi2026wsp}. Early studies apply WSP to small models on mathematics and question answering, and they find that edits remove obstacles more easily than they add abilities \citep{takehi2026wsp,hu2026exploring,khamis2026aha,ok2026wsp}. Whether such edits can improve a pretrained reasoning model on a hard generation task has received little study.

RTL generation is a natural test for this question, because every candidate design can be checked by simulation. A set of practice specifications with reference testbenches gives an exact signal for each proposed edit. At inference time, many candidate designs can be compared by their simulated behaviour without any reference solution.

We present \sys{}, an agentic framework that improves open models for RTL generation without gradient training (\cref{fig:overview}). Its first part is neuron programming, a new form of WSP that we introduce. We look for recurring failure habits of the model on practice tasks, write one neuron for each habit and keep the edit only when a paired testbench comparison on fresh practice tasks favours it. The method is gradient-free in the sense that no optimizer updates any weight. It still uses practice data, for diagnosis and for acceptance tests. At inference time, independent agents propose designs with compile and edit tools, and a label-free selection stage delivers one design per task. The selection groups the candidates by simulated behaviour and resolves disagreements with long simulations, judgements by the same model and the input-output examples stated in the specification. We build two systems, \sys{} on a 27B model and \sysmini{} on a 4B model.

Our contributions are as follows:
\begin{itemize}
\item \textbf{Neuron programming.} We introduce neuron programming, a new gradient-free form of WSP. Each program is a single neuron whose trigger, read and write vectors are set in closed form to remove a failure habit identified on practice tasks, and an edit is accepted only after a paired statistical test. We show that an inactive program perturbs the model by an exponentially small amount, and that the accuracy gain of an edit factorizes into how often it changes the output and how often such a change helps.
\item \textbf{A label-free agentic framework for RTL generation.} Tool-using agents propose candidate designs, and a selection stage delivers one of them without any testbench or reference design. Candidates are grouped by simulated behaviour, and disagreements are resolved through long simulations, same-model judgement and the examples stated in the specification.
\item \textbf{Strong results with open models.} With Qwen3.8-27B on a single 32\,GB GPU, \sys{} reaches 94.9\% pass@1 on VerilogEval v2 and 92.0\% on RTLLM v2.1, above every published system in our comparison. On Qwen3.5-4B, the weight edits alone add 23.7 points, and the full framework reaches 77.6\%.
\item \textbf{Transfer beyond hardware.} With the same weights and a Python port of its harness, the 4B system also improves Python code generation and solves 63.4\% of recent LiveCodeBench v6 problems, against 55.8\% reported for its base model. The weight edits alone, built only on RTL practice tasks, raise Python pass@1 of the 4B model from 67.7\% to 77.5\% on MBPP+ and from 84.8\% to 91.5\% on HumanEval+.
\end{itemize}

\section{Related Work}

VerilogEval \citep{liu2023verilogeval,pinckney2025revisiting} and RTLLM \citep{lu2024rtllm,liu2024openllmrtl} established specification-to-RTL benchmarks, and ChipBench \citep{yu2026chipbench} and CVDP \citep{pinckney2025cvdp} added harder and larger tasks. Open models improve mainly through fine-tuning on synthesized or filtered corpora \citep{liu2024rtlcoder,cui2024origen,liu2025craftrtl}, reinforcement learning with testbench rewards \citep{zhu2025codevr1,yu2026chipmate} or guided decoding \citep{pei2024betterv}. Agentic systems add compilation, simulation and debugging loops \citep{zhao2025mage,ho2025verilogcoder}. Our framework uses practice tasks only for diagnosis and acceptance tests, and its edits change fewer than 0.005\% of the weights of the 4B model.

Knowledge editing locates and rewrites facts in the multilayer perceptron (MLP) layers of a transformer \citep{meng2022locating,meng2023memit,mitchell2022fast,fang2025alphaedit}, task arithmetic adds weight differences between fine-tuned models \citep{ilharco2023editing}, and low-rank adaptation is the standard low-cost form of fine-tuning \citep{hu2022lora}. Activation steering adds directions to the residual stream at inference time \citep{turner2023activation,zou2023representation,rimsky2024steering,arditi2024refusal}. Closer to our setting, MeG adds one neuron whose weights a diffusion model generates for each edit \citep{wan2026meg}, and Neuron Patching repairs code models by updating a few neurons chosen for given wrong outputs \citep{gu2023neuronpatching}. Most of these methods fit edits by optimization, generate them with a learned model or apply them at every position. Our programs are single neurons with an explicit trigger, built in closed form, following the key-value view of MLP neurons \citep{geva2021transformer,geva2022transformer}. Hand-written weights have also been compiled into transformers for small programs \citep{weiss2021thinking,lindner2023tracr}, an association can be installed into one spare unit with key and value read from the weights \citep{oskin2026looking}, and zero-weight neurons preserve the function of a network \citep{chen2016net2net}.

WSP was introduced as an LLM that edits the weights of a target model with forward passes only \citep{takehi2026wsp}. Later work studies which capabilities such edits reach and how they transfer \citep{hu2026exploring}, writes edits without an LLM in the loop \citep{khamis2026aha} and lets LLMs design architectures that are easier to write \citep{ok2026wsp}. WSP has also written small medical image classifiers without training data \citep{lindholz2026medical} and repaired robot policies through edits of embedding rows, evaluated with an exact paired test \citep{zhang2026robots}, and policies written by WSP have been verified by probabilistic model checking inside the writing loop \citep{gross2026wspmc}. Whether LLMs reason about the weights they read remains debated \citep{ahmed2026reason}. Neuron programming combines three elements that these works do not combine. Its unit of edit is a single neuron with an explicit trigger, written in closed form. Every edit must pass a paired testbench comparison, and the accepted stack is confirmed by a test with a bounded type-I error. Its targets are pretrained reasoning models with 4B and 27B parameters on a generation task, and the edited model feeds test-time selection.

Self-consistency votes over sampled answers \citep{wang2023selfconsistency}, AlphaCode clusters programs by their outputs on generated inputs \citep{li2022alphacode}, and CodeT and execution-based minimum Bayes risk decoding score programs with generated tests \citep{chen2023codet,shi2022natural}. Our selection groups hardware designs by cycle-accurate simulation and adds long simulations, same-model judgement and replay of the examples stated in the specification. Our agents follow the reason-and-act pattern of tool-using LLMs \citep{yao2023react}.

\section{Neuron Programming}
\label{sec:wsp}

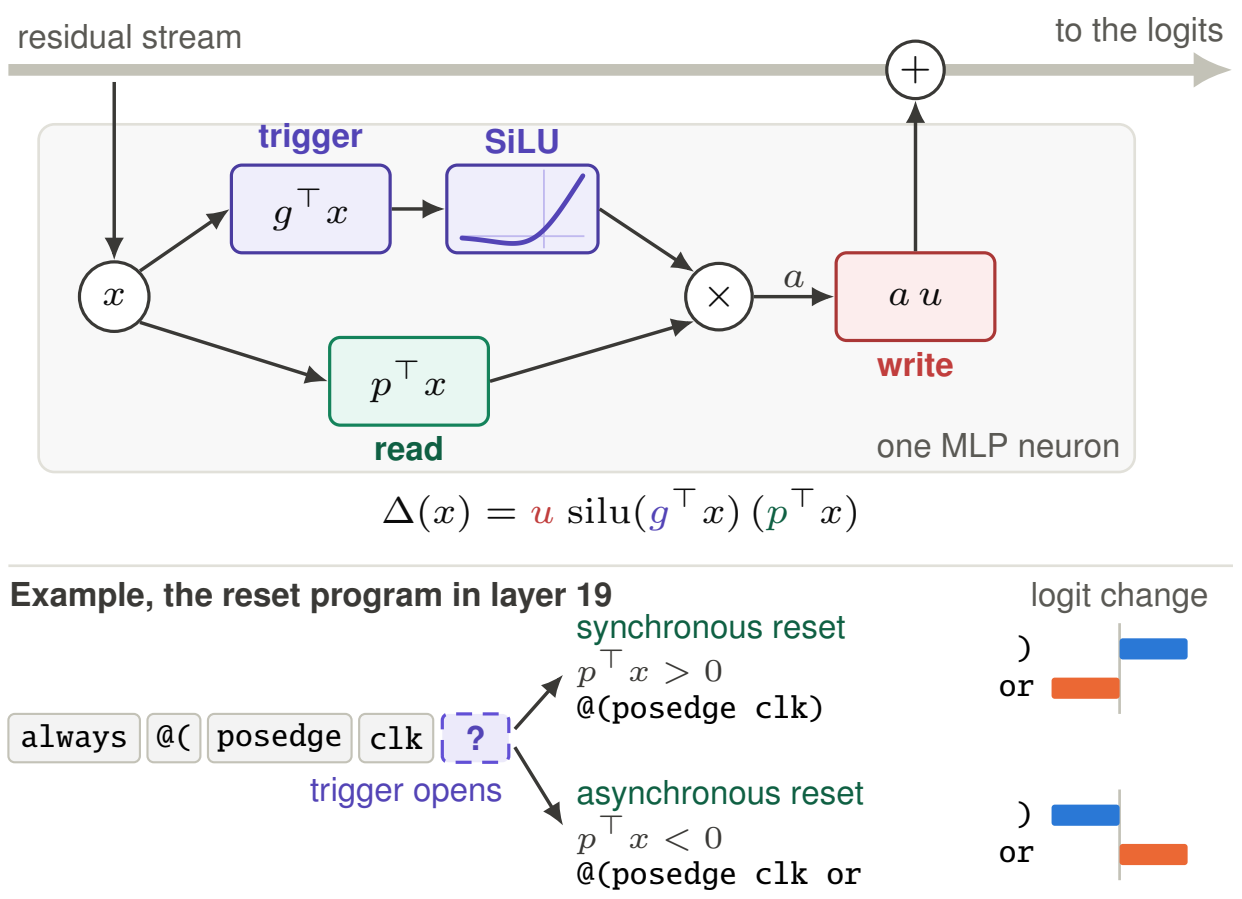
\begin{figure}[t]
\centering
\begin{tikzpicture}[x=1cm,y=1cm,
    op/.style={draw=#1!75!black, fill=#1!10, line width=0.6pt, rounded corners=2pt, minimum height=0.58cm,
               minimum width=1.05cm, inner sep=2pt, font=\figfont, text=kink},
    node0/.style={circle, draw=kinkb, fill=white, line width=0.6pt, inner sep=0pt, font=\figfont, text=kink},
    arr/.style={-{Latex[length=1.5mm,width=1.3mm]}, line width=0.65pt, draw=kinkb},
    role/.style={font=\figfontS\bfseries, inner sep=1.6pt}]
  \draw[kaxis, line width=2.2pt, -{Latex[length=2.6mm,width=2.4mm]}] (0.05,5.78) -- (8.15,5.78);
  \node[font=\figfontS, text=kmuted, anchor=south west, inner sep=1.5pt] at (0.05,5.86) {residual stream};
  \node[font=\figfontS, text=kmuted, anchor=south east, inner sep=1.5pt] at (8.15,5.86) {to the logits};
  \path[fill=kink!3, draw=kgrid, line width=0.5pt, rounded corners=3pt] (0.25,3.12) rectangle (7.5,5.42);
  \node[font=\figfontS, text=kmuted, anchor=south east, inner sep=2.5pt] at (7.5,3.12) {one MLP neuron};
  \draw[arr] (0.75,5.7) -- (0.75,4.52);
  \node[node0, minimum size=0.46cm] (x) at (0.75,4.28) {$x$};
  \node[op=kviolet] (g) at (2.05,4.86) {$g\T x$};
  \node[role, text=kviolet!85!black, anchor=south] at (g.north) {trigger};
  \node[op=kviolet, minimum width=1.0cm] (s) at (3.45,4.86) {};
  \node[role, text=kviolet!85!black, anchor=south] at (s.north) {SiLU};
  \draw[kviolet!40, line width=0.35pt] (3.04,4.68) -- (3.86,4.68);
  \draw[kviolet!40, line width=0.35pt] (3.5923,4.6) -- (3.5923,5.12);
  \draw[kviolet!85!black, line width=0.9pt, line cap=round] plot[domain=-4:2.5, samples=80, smooth]
    ({3.05+(\x+4)*0.1231}, {4.68+0.173*\x/(1+exp(-\x))});
  \node[op=kaqua] (p) at (2.7,3.72) {$p\T x$};
  \node[role, text=kaqua!55!black, anchor=north] at (p.south) {read};
  \node[node0, minimum size=0.44cm] (m) at (4.75,4.28) {$\times$};
  \node[op=kred] (u) at (6.05,4.28) {$a\,u$};
  \node[role, text=kred!85!black, anchor=north] at (u.south) {write};
  \node[node0, minimum size=0.38cm] (plus) at (6.05,5.78) {$+$};
  \draw[arr] (x.north east) -- (g.west);
  \draw[arr] (x.south east) -- (p.west);
  \draw[arr] (g.east) -- (s.west);
  \draw[arr] (s.east) -- (m.north west);
  \draw[arr] (p.east) -- (m.south west);
  \draw[arr] (m.east) -- node[above, font=\figfont, text=kinkb, inner sep=1.5pt] {$a$} (u.west);
  \draw[arr] (u.north) -- (plus.south);
  \node[font=\figfont, text=kink, anchor=north, inner sep=0pt] at (4.1,3.02)
    {$\Delta(x) = {\color{kred!85!black}u}\;\mathrm{silu}({\color{kviolet!85!black}g}\T x)\,({\color{kaqua!55!black}p}\T x)$};
  \draw[kgrid, line width=0.5pt] (0.05,2.5) -- (8.15,2.5);
  \node[font=\figfontS\bfseries, text=kinkb, anchor=north west, inner sep=0pt] at (0.05,2.4) {Example, the reset program in layer 19};
  \node[font=\figfontS, text=kmuted, anchor=north, inner sep=0pt] at (7.4,2.4) {logit change};
  \foreach \t/\xa/\xb in {always/0.05/0.92, @(/0.97/1.32, posedge/1.37/2.32, clk/2.37/2.86} {
    \path[fill=kink!5, draw=kaxis, line width=0.4pt, rounded corners=1.2pt] (\xa,1.2) rectangle (\xb,1.52);
    \node[anchor=center, inner sep=0pt] at ($(\xa,1.36)!0.5!(\xb,1.36)$) {\figtt{\t}};
  }
  \path[fill=kviolet!12, draw=kviolet, line width=0.6pt, densely dashed, rounded corners=1.2pt] (2.92,1.2) rectangle (3.36,1.52);
  \node[font=\figfontS\bfseries, text=kviolet!85!black, anchor=center, inner sep=0pt] at (3.14,1.36) {?};
  \node[font=\figfontS, text=kviolet!85!black, anchor=north east, inner sep=1.5pt] at (3.38,1.16) {trigger opens};
  \draw[arr] (3.4,1.42) -- (3.72,1.78);
  \draw[arr] (3.4,1.3) -- (3.72,0.78);
  \foreach \yy/\spec/\sgn/\cA/\cB/\code in {1.82/synchronous/{>}/1/-1/{@(posedge clk)}, 0.72/asynchronous/{<}/-1/1/{@(posedge clk or}} {
    \node[font=\figfontS, text=kaqua!55!black, anchor=west, inner sep=0pt] at (3.8,\yy+0.26) {\spec\ reset};
    \node[font=\figfontS, text=kinkb, anchor=west, inner sep=0pt] at (3.8,\yy) {$p\T x \sgn 0$};
    \node[anchor=west, inner sep=0pt] at (3.8,\yy-0.27) {\figtt{\code}};
    \draw[kaxis, line width=0.5pt] (7.4,\yy-0.3) -- (7.4,\yy+0.3);
    \node[anchor=east, inner sep=1pt] at (6.88,\yy+0.13) {\figtt{)}};
    \node[anchor=east, inner sep=1pt] at (6.88,\yy-0.13) {\figtt{or}};
    \path[fill=kblue, rounded corners=0.6pt] (7.4,\yy+0.06) rectangle ({7.4+0.45*\cA},\yy+0.2);
    \path[fill=korange, rounded corners=0.6pt] (7.4,\yy-0.2) rectangle ({7.4+0.45*\cB},\yy-0.06);
  }
\end{tikzpicture}
\caption{A program is one MLP neuron with three hand-written vectors. The trigger opens only where a habit starts, the read measures a signal that the model computes and the write moves the logits of chosen tokens. In the reset example, the sign of $p\T x$ decides between the two possible next tokens.}
\label{fig:program}
\end{figure}

\subsection{Programs as neurons}

Both models use SwiGLU MLPs \citep{shazeer2020glu}. Neuron $i$ of an MLP adds the vector
\begin{equation}
  \Delta_i(x) = u_i\,\silu(g_i\T x)\,(p_i\T x)
  \label{eq:neuron}
\end{equation}
to the residual stream, where $x$ is the normalized input of the MLP, $g_i$ and $p_i$ are the $i$-th rows of the gate and up projections, $u_i$ is the $i$-th column of the down projection and $\silu(z)=z\,\sigma(z)$. A program is a choice of the three vectors of one neuron (\cref{fig:program}). The trigger $g$ decides where the neuron is active, the read vector $p$ decides what it measures there and the write vector $u$ decides which token logits it moves. A pair of neurons is used when a stronger or a saturating push is needed.

Let $X_c$ hold the states where a habit $c$ starts, for example after the keywords \texttt{output reg} in a port list, and $X_0$ states from ordinary positions. The trigger is
\begin{equation}
  g = \gamma\,(d - \tau b), \qquad d = \bar{x}_c - \bar{x}_0,
  \label{eq:trigger}
\end{equation}
where $\bar{x}_c$ and $\bar{x}_0$ are the means of $X_c$ and $X_0$, the threshold $\tau$ lies above a high quantile of $d\T x$ on $X_0$ and $\gamma$ is a gain. A trigger is kept only if it also separates held-out states. A neuron has no bias, so the threshold is carried by a direction $b$ whose projection $b\T x$ is nearly constant and normalized to about 1. The read vector $p$ measures a signal that the model already computes, such as the reset wording of the specification, or equals $b$ for a constant push. Where a last-layer program fires with activation $a = \silu(g\T x)\,(p\T x)$, the logits change by
\begin{equation}
  z' - z = a\, J u + O\big(a^2 \|u\|^2\big),
  \label{eq:dose}
\end{equation}
where $J$ is the Jacobian of the final normalization followed by the unembedding, averaged over the captured states. Programs in earlier layers use the same $J$ as an approximation. Let $J_T$ and $J_P$ hold the rows of $J$ for the target tokens $T$ and for a protected set $P$, which contains the special tokens of the chat template and the current top tokens of the captured states. The write vector is the shortest vector that moves the target logits by $\delta$ per unit activation and leaves the protected logits unchanged. With $A$ stacking $J_T$ over $J_P$, it has the closed form
\begin{equation}
  u = \operatorname*{arg\,min}_{J_T u = \delta,\; J_P u = 0} \|u\|
    = A\T \big(A A\T\big)^{-1} \begin{pmatrix} \delta \\ 0 \end{pmatrix}.
  \label{eq:write}
\end{equation}

For Qwen3.5-4B we overwrite neurons of layers 19 to 31 that have low mean activation on practice traces, so every paired comparison measures the joint effect of removing such a neuron and adding the program. A neuron of this model holds 7,680 weights, and every weight outside the edited neurons stays unchanged. For the 27B model we append 128 neurons with zero weights to every MLP, which widens it from 17,408 to 17,536 neurons. The spare neurons are stored apart from the original weights, so the unedited model is reproduced exactly, and programs are written only into them.

\subsection{Idle programs}

The negative lobe of SiLU reaches about $-0.28$ near $z=-1.28$, so a program leaks a little when its trigger is nearly open. The leak vanishes exponentially once the trigger input lies far below zero. Proofs of the three propositions are given in \cref{app:proofs}.

\begin{proposition}[Off-trigger bound]\label{prop:closed}
If $g\T x \le -\tau$ for some $\tau \ge 1$, then $\|\Delta(x)\| \le \tau e^{-\tau}\,|p\T x|\,\|u\|$.
\end{proposition}

We therefore set the gain so that ordinary states fall far below the threshold, and we count firings on several hundred thousand ordinary positions before any paired test. One early edit with a shallow trigger leaked into every position through the final normalization and was discarded.

\begin{algorithm}[t]
\caption{Neuron programming}
\label{alg:wsp}
\small
\begin{algorithmic}[1]
\REQUIRE model $M$, practice tasks $\mathcal{P}$ with testbenches
\ENSURE edited model $M$
\FOR{each frequent failure class $c$ of $M$ on $\mathcal{P}$}
  \STATE $X_c, X_0 \gets$ states where $c$ starts, ordinary states
  \IF{no linear trigger separates $X_c$ from $X_0$}
    \STATE \textbf{continue} \acmt{not fixable by one neuron}
  \ENDIF
  \STATE $g \gets \gamma\,(\bar{x}_c - \bar{x}_0 - \tau b)$ \acmt{Eq.~\eqref{eq:trigger}}
  \STATE $u \gets A\T (A A\T)^{-1} (\delta, 0)$ \acmt{Eq.~\eqref{eq:write}}
  \STATE $M' \gets M$ with one neuron set to $(g, p, u)$
  \STATE $(n_{10}, n_{01}) \gets$ paired run of $M'$ and $M$ on fresh tasks
  \IF{$c$ is removed, no new class appears and $n_{10} > n_{01}$}
    \STATE $M \gets M'$
  \ENDIF
\ENDFOR
\STATE confirm the stack with \cref{prop:accept}
\STATE \textbf{return} $M$
\end{algorithmic}
\end{algorithm}

\subsection{What an edit can gain}

Let the unedited and the edited model solve the same task with the same sampling seeds, and let $Y_0, Y_1 \in \{0,1\}$ indicate whether their designs pass. Let $D$ be the event that the two generated token sequences differ.

\begin{proposition}[Rescue identity]\label{prop:rescue}
$\E[Y_1] - \E[Y_0] = \Prob(D)\,(r - s)$, where $r = \Prob(Y_0=0, Y_1=1 \mid D)$ and $s = \Prob(Y_0=1, Y_1=0 \mid D)$. If idle programs change no sampled token, then $D \subseteq F$ for the event $F$ that some trigger opens during the generation, and hence $|\E[Y_1] - \E[Y_0]| \le \Prob(F)$.
\end{proposition}

A fix for a rare habit therefore has a small ceiling however well it works. A change of output also redraws every later token, which turns some passes into failures by chance and keeps $r - s$ below the rate at which the targeted error is removed.

\subsection{Accepting an edit}

An edit is tested on practice tasks that were not used to build it. Both models run with the same seeds, and we count the tasks $n_{10}$ that only the edited model solves and the tasks $n_{01}$ that only the unedited model solves.

\begin{proposition}[Paired acceptance]\label{prop:accept}
Let tasks be independent and $m = n_{10} + n_{01}$. If $r \le s$, then $\Prob(n_{10} \ge k \mid m) \le \Prob(\mathrm{Bin}(m, 1/2) \ge k)$ for every $k$. Rejecting $r \le s$ when the binomial tail at the observed $n_{10}$ is at most $\alpha$ therefore has a type-I error of at most $\alpha$.
\end{proposition}

By \cref{prop:rescue}, a rejection implies a positive accuracy gain on the practice distribution. The test is the exact McNemar test \citep{mcnemar1947note}. Single edits were screened with paired runs of 128 to 512 practice tasks, which detect only large effects. We kept an edit when it removed its target failure class, created no new class in the failure reports and had $n_{10} > n_{01}$. Stacks of edits were then confirmed with \cref{prop:accept} on the full practice pool. \cref{alg:wsp} summarizes the procedure. In our runs Claude, an LLM from Anthropic, proposed every program under the supervision of the authors, through a restricted kernel that forbids gradients, optimizers and fitted parameters and never edits the embeddings, the unembedding or the final normalization.

\subsection{The edits}
\label{sec:edits}

The edits of Qwen3.5-4B occupy 25 neurons with 192,000 weights, listed in \cref{app:edits}. In 48 of these weights the new value rounds to the old one, so 191,952 weights change. Sixteen act on the model alone. They end runaway reasoning, remove six syntax habits, such as decimal digits after a binary base and ports declared twice, and make the sensitivity list follow the reset wording of the specification. The other nine shape agent turns, for example by opening the first answer with a tool call. The 27B model shows far fewer such habits and carries one such tool-call program in two spare neurons of its last layer, with accuracy on practice tasks unchanged within noise.

\section{The Agentic Framework}
\label{sec:selection}

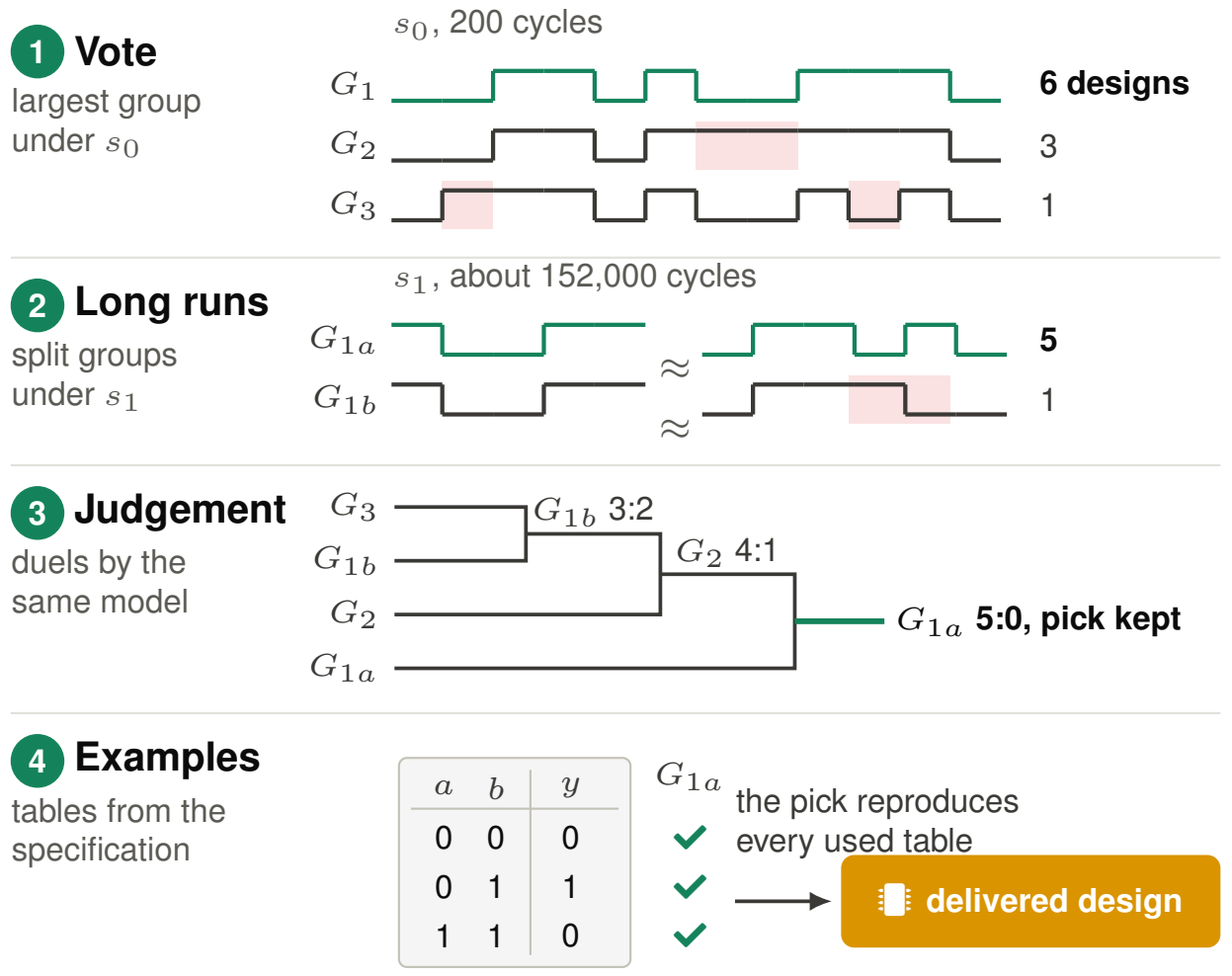
\begin{figure}[t]
\centering
\begin{tikzpicture}[x=1cm,y=1cm,
    arr/.style={-{Latex[length=1.5mm,width=1.3mm]}, line width=0.7pt, draw=kinkb},
    glab/.style={font=\figfontS, text=kinkb, anchor=east, inner sep=1.5pt},
    gsz/.style={font=\figfontS\bfseries, text=kink, anchor=west, inner sep=1.5pt}]
  \def\cw{0.34}
  \def\amp{0.2}
  \newcommand{\wave}[4]{%
    \gdef\prevbit{-1}%
    \foreach \b [count=\i from 0] in {#3} {
      \pgfmathsetmacro{\xa}{#1+\i*\cw}
      \pgfmathsetmacro{\xb}{#1+(\i+1)*\cw}
      \pgfmathsetmacro{\yl}{#2+\b*\amp}
      \draw[#4, line width=0.8pt] (\xa,\yl) -- (\xb,\yl);
      \ifnum\prevbit>-1 \ifnum\prevbit=\b\relax\else \draw[#4, line width=0.8pt] (\xa,#2) -- (\xa,#2+\amp); \fi\fi
      \xdef\prevbit{\b}
    }}
  \newcommand{\miss}[3]{\foreach \i in {#3} { \path[fill=kred!16] ({#1+\i*\cw},#2-0.06) rectangle ({#1+(\i+1)*\cw},#2+\amp+0.06); }}
  \newcommand{\stage}[4]{%
    \node[circle, fill=kaqua!75!black, text=white, font=\figfontS\bfseries, minimum size=0.36cm, inner sep=0pt] at (0.18,#1) {#2};
    \node[font=\figfontT, text=kink, anchor=west, inner sep=0pt] at (0.42,#1) {#3};
    \node[font=\figfontS, text=kmuted, anchor=north west, inner sep=0pt, align=left] at (0.0,#1-0.24) {#4};}
  \stage{6.55}{1}{Vote}{largest group\\under $s_0$}
  \node[font=\figfontS, text=kmuted, anchor=south west, inner sep=0pt] at (2.55,6.62) {$s_0$, 200 cycles};
  \miss{2.55}{5.82}{6,7}
  \miss{2.55}{5.42}{1,9}
  \wave{2.55}{6.22}{0,0,1,1,0,1,0,0,1,1,1,0}{kaqua!75!black}
  \wave{2.55}{5.82}{0,0,1,1,0,1,1,1,1,1,1,0}{kinkb}
  \wave{2.55}{5.42}{0,1,1,1,0,1,0,0,1,0,1,0}{kinkb}
  \node[glab] at (2.53,6.32) {$G_1$};
  \node[glab] at (2.53,5.92) {$G_2$};
  \node[glab] at (2.53,5.52) {$G_3$};
  \node[gsz] at (6.83,6.32) {6 designs};
  \node[gsz, font=\figfontS, text=kinkb] at (6.83,5.92) {3};
  \node[gsz, font=\figfontS, text=kinkb] at (6.83,5.52) {1};
  \draw[kgrid, line width=0.4pt] (0,5.17) -- (8.1,5.17);
  \stage{4.85}{2}{Long runs}{split groups\\under $s_1$}
  \node[font=\figfontS, text=kmuted, anchor=south west, inner sep=0pt] at (2.55,4.92) {$s_1$, about 152,000 cycles};
  \miss{2.55}{4.12}{9,10}
  \wave{2.55}{4.52}{1,0,0,1,1}{kaqua!75!black}
  \wave{2.55}{4.12}{1,0,0,1,1}{kinkb}
  \wave{4.63}{4.52}{0,1,1,0,1,0}{kaqua!75!black}
  \wave{4.63}{4.12}{0,1,1,1,0,0}{kinkb}
  \node[font=\figfont, text=kmuted] at (4.45,4.42) {$\approx$};
  \node[font=\figfont, text=kmuted] at (4.45,4.02) {$\approx$};
  \node[glab] at (2.53,4.62) {$G_{1a}$};
  \node[glab] at (2.53,4.22) {$G_{1b}$};
  \node[gsz] at (6.83,4.62) {5};
  \node[gsz, font=\figfontS, text=kinkb] at (6.83,4.22) {1};
  \draw[kgrid, line width=0.4pt] (0,3.78) -- (8.1,3.78);
  \stage{3.46}{3}{Judgement}{duels by the\\same model}
  \foreach \n/\yy in {$G_3$/3.5, $G_{1b}$/3.14, $G_2$/2.78, $G_{1a}$/2.42} { \node[glab] at (2.53,\yy) {\n}; }
  \draw[kinkb, line width=0.7pt] (2.57,3.5) -- (3.45,3.5) -- (3.45,3.14) -- (2.57,3.14);
  \draw[kinkb, line width=0.7pt] (3.45,3.32) -- (4.35,3.32) -- (4.35,2.78) -- (2.57,2.78);
  \draw[kinkb, line width=0.7pt] (4.35,3.05) -- (5.25,3.05) -- (5.25,2.42) -- (2.57,2.42);
  \draw[kaqua!75!black, line width=1.1pt] (5.25,2.735) -- (5.85,2.735);
  \node[font=\figfontS, text=kinkb, anchor=south, inner sep=1.2pt] at (3.9,3.32) {$G_{1b}$ 3:2};
  \node[font=\figfontS, text=kinkb, anchor=south, inner sep=1.2pt] at (4.8,3.05) {$G_2$ 4:1};
  \node[font=\figfontS\bfseries, text=kink, anchor=west, inner sep=2pt] at (5.85,2.735) {$G_{1a}$ 5:0, pick kept};
  \draw[kgrid, line width=0.4pt] (0,2.12) -- (8.1,2.12);
  \stage{1.8}{4}{Examples}{tables from the\\specification}
  \begin{scope}[shift={(2.6,0)}]
    \path[fill=kink!4, draw=kaxis, line width=0.4pt, rounded corners=1.5pt] (0,0.42) rectangle (1.55,1.82);
    \node[font=\figfontS\bfseries, text=kinkb] at (0.3,1.62) {$a$};
    \node[font=\figfontS\bfseries, text=kinkb] at (0.65,1.62) {$b$};
    \node[font=\figfontS\bfseries, text=kinkb] at (1.15,1.62) {$y$};
    \draw[kaxis, line width=0.4pt] (0.08,1.48) -- (1.47,1.48);
    \draw[kaxis, line width=0.4pt] (0.88,0.5) -- (0.88,1.74);
    \foreach \a/\bb/\y [count=\r] in {0/0/0, 0/1/1, 1/1/0} {
      \pgfmathsetmacro{\yy}{1.62-0.33*\r}
      \node[font=\figfontS, text=kink] at (0.3,\yy) {\a};
      \node[font=\figfontS, text=kink] at (0.65,\yy) {\bb};
      \node[font=\figfontS, text=kink] at (1.15,\yy) {\y};
      \node[font=\figfontS, text=kaqua!75!black] at (1.95,\yy) {\faCheck};
    }
    \node[font=\figfontS, text=kinkb, anchor=south] at (1.95,1.48) {$G_{1a}$};
  \end{scope}
  \node[font=\figfontS, text=kinkb, anchor=west, align=left, inner sep=0pt] at (4.85,1.38) {the pick reproduces\\every used table};
  \draw[arr] (4.85,0.86) -- (5.5,0.86);
  \path[fill=kyellow!92!black, rounded corners=2.5pt] (5.56,0.55) rectangle (8.1,1.17);
  \node[text=white, font=\figfontS\bfseries, anchor=center] at (6.83,0.86) {{\faMicrochip}\hspace{1mm}delivered design};
\end{tikzpicture}
\caption{Label-free selection on a toy task with ten candidates. (1) Equal outputs under $s_0$ form groups, and the vote takes the largest. (2) Long runs split groups that agree only on short runs. (3) For \sys{}, a knockout of same-model duels starts from the smallest group, with group sizes hidden. (4) Tables copied from the specification can overturn a pick that contradicts them.}
\label{fig:selection}
\end{figure}

\paragraph{Agents.} For each task, $N$ agents with independent seeds work on the same specification. Each agent receives the task text of the benchmark, followed by a few short cards of standard design practice chosen by keywords of the specification. It can compile its design with Icarus Verilog \citep{williams2026icarus}, edit it and submit it. After a successful compile the harness also reports lint warnings, outputs that stay unknown on generated inputs, ports of the specification that the design lacks and contradictions with plain statements of the specification, such as the reset kind and the port widths. A reasoning runaway or an overlong conversation is restarted in a fresh context that carries a summary. The first design an agent submits is its candidate. \cref{app:harness} lists the harness features.

\paragraph{Grouping and vote.} Selection runs in four stages (\cref{fig:selection}). Let $d_1, \dots, d_N$ be the candidates of a task and $y(d, s)$ the outputs of design $d$, cycle by cycle, under a stimulus $s$. Designs with $y(d, s) = y(d', s)$ form a group. The vote returns the largest group under a shared stimulus $s_0$ of 200 cycles with random inputs and a reset about every 16 cycles. Ties go to groups with known outputs and to designs that declare the ports named in the specification.

\paragraph{Long simulations.} Counters, timers and long input sequences rarely show under $s_0$. A long stimulus $s_1$ of about 152,000 cycles mixes random, slowly changing and held inputs with rare resets, and it covers every input combination of a combinational design with up to 14 input bits. Inside each group, the representative becomes a member of the largest sub-group that still agrees under $s_1$. For \sys{} a third stimulus $s_2$ holds each 1-bit input for $K-2, \dots, K+2$ cycles, for every constant $2 \le K \le 1024$ that occurs in a candidate or in the specification, which exposes designs that are wrong by one cycle at a threshold. Groups that split under these runs are replaced by their sub-groups, ordered by size.

\paragraph{Same-model judgement.} For \sys{} the groups meet in a knockout that starts from the smallest group. In each duel the model reads the specification and one design from each group, with comments removed and group sizes hidden, and five votes decide. A champion that differs from the current pick replaces it only if a second query confirms it. In that query five respondents see one design per group, and the champion's group must be named at least twice and more often than any other. The 4B model judged code less reliably on practice tasks, so \sysmini{} asks a behaviour question instead. The model reads the specification and the shortest simulation run on which two groups differ, without any code. The pick moves only on a unanimous three-vote verdict, or else on a unanimous three-vote code duel.

\paragraph{Example replay.} Many specifications state examples such as truth-table rows, Karnaugh-map cells and waveforms. For \sys{} the model copies them into tables in six samples, three with thinking and three without, and a fixed harness replays every table $T$ on every candidate. We write $d \models T$ when design $d$ reproduces $T$. A table is used if at least two samples give it and $d \models T$ for some candidate $d$. A pick that reproduces no used table is replaced by the first candidate in group order that reproduces all of them.

\paragraph{Protocol.} Every judgement, duel and table comes from the same model as the agents, and no agent sees the benchmark's testbench, a reference design or a score. The vote, the long-run step and the judgement rules were fixed before their benchmark runs. The last stage of \sys{}, which adds the constant-directed stimulus and the example replay, was developed with VerilogEval v2 outputs in view, so VerilogEval is a development set for that stage. RTLLM v2.1, ChipBench and CVDP were run once with the frozen system.

\section{Experimental Setup}

\begin{figure*}[t]
\centering
\begin{tikzpicture}[x=1cm,y=1cm]
  \newcommand{\cell}[5]{%
    \ifdim #3 pt<0pt
      \path[draw=kred!70, densely dashed, line width=0.5pt, rounded corners=0.8pt] (#1,#2-0.09) rectangle (#1+0.95,#2+0.09);
      \node[anchor=west, inner sep=0pt, font=\figfontS, text=kred!80!black] at (#1+1.02,#2) {TBD};
    \else
      \pgfmathsetmacro{\w}{#3*#5}
      \path[kind#4, rounded corners=0.8pt] (#1,#2-0.09) rectangle (#1+\w,#2+0.09);
      \node[anchor=west, inner sep=0pt, val#4] at (#1+\w+0.06,#2) {#3};
    \fi}
  \newcommand{\colhead}[7]{%
    \node[anchor=south west, inner sep=0pt] at (#1,#2) {\figfontT #3\hspace{1.2mm}{\figfontS\mdseries\color{kmuted}#4}};
    \foreach \t in {25,50,75} { \draw[kgrid, line width=0.4pt] (#1+\t*#7,#5) -- (#1+\t*#7,#6); }
    \draw[kaxis, line width=0.5pt] (#1,#5) -- (#1,#6);}
  \def\dy{0.27}    %
  \def\sc{0.030}   %
  \def\cw{3.62}    %
  \def\nr{17}      %
  \pgfmathsetmacro{\bot}{0.02-\dy*\nr}
  \colhead{0}{0.2}{VerilogEval v2}{156 tasks}{0.06}{\bot}{\sc}
  \colhead{\cw}{0.2}{RTLLM v2.1}{50 designs}{0.06}{\bot}{\sc}
  \colhead{2*\cw}{0.2}{ChipBench}{30 tasks}{0.06}{\bot}{\sc}
  \colhead{3*\cw}{0.2}{CVDP cid003}{78 problems}{0.06}{\bot}{\sc}
  \foreach \n/\a/\b/\c/\d/\k [count=\i] in {
      \sys{}/94.9/92.0/66.7/59.0/A,
      Claude Opus 4.7/86.9/64.8/31.3/42.8/P,
      GPT-5.5/84.7/63.2/30.7/44.0/P,
      ChipMATE-Agents-9B/80.1/75.8/36.7/40.4/P,
      \sysmini{}/77.6/64.0/30.0/32.1/B,
      DeepSeek-R1/77.5/64.7/26.7/27.7/P,
      ChipMATE-Verilog-9B/75.3/71.9/30.0/28.1/P,
      ChipMATE-Agents-4B/75.0/74.6/33.3/32.1/P,
      Qwen3.8-27B (base model)/73.1/66.0/56.7/37.2/a,
      CodeV-R1/68.8/68.0/30.0/26.8/P,
      DeepSeek-Coder/68.5/57.6/16.7/22.3/P,
      ChipMATE-Verilog-4B/67.4/68.0/26.7/24.7/P,
      DeepSeek-V4/67.3/58.8/18.0/21.5/P,
      CodeV-R1 (distilled)/65.2/57.2/13.3/26.2/P,
      GPT-4o/64.1/56.5/20.0/39.0/P,
      Qwen3.5-9B/48.5/36.1/13.3/13.3/P,
      Qwen3.5-4B (base model)/31.7/19.0/11.7/9.0/b} {
    \pgfmathsetmacro{\y}{0.02-\dy*(\i-0.5)}
    \node[anchor=east, inner sep=0pt, name\k] at (-0.08,\y) {\n};
    \cell{0}{\y}{\a}{\k}{\sc}
    \cell{\cw}{\y}{\b}{\k}{\sc}
    \cell{2*\cw}{\y}{\c}{\k}{\sc}
    \cell{3*\cw}{\y}{\d}{\k}{\sc}
  }
  \begin{scope}[shift={(-2.55,0.9)}]
    \path[kindA, rounded corners=0.8pt] (0,-0.09) rectangle (0.42,0.09);
    \node[anchor=west, font=\figfontS, text=kink] at (0.47,0) {\sys{}};
    \path[kindB, rounded corners=0.8pt] (1.75,-0.09) rectangle (2.17,0.09);
    \node[anchor=west, font=\figfontS, text=kink] at (2.22,0) {\sysmini{}};
    \path[kinda, rounded corners=0.8pt] (4.15,-0.09) rectangle (4.36,0.09);
    \path[kindb, rounded corners=0.8pt] (4.4,-0.09) rectangle (4.61,0.09);
    \node[anchor=west, font=\figfontS, text=kink] at (4.66,0) {their base models in our runs, one answer without tools};
    \path[kindP, rounded corners=0.8pt] (10.6,-0.09) rectangle (11.02,0.09);
    \node[anchor=west, font=\figfontS, text=kink] at (11.07,0) {published score};
  \end{scope}
\end{tikzpicture}
\caption{Pass@1 (\%) of our systems, their base models and published systems \citep{yu2026chipmate} on four RTL benchmarks.}
\label{fig:compare}
\end{figure*}
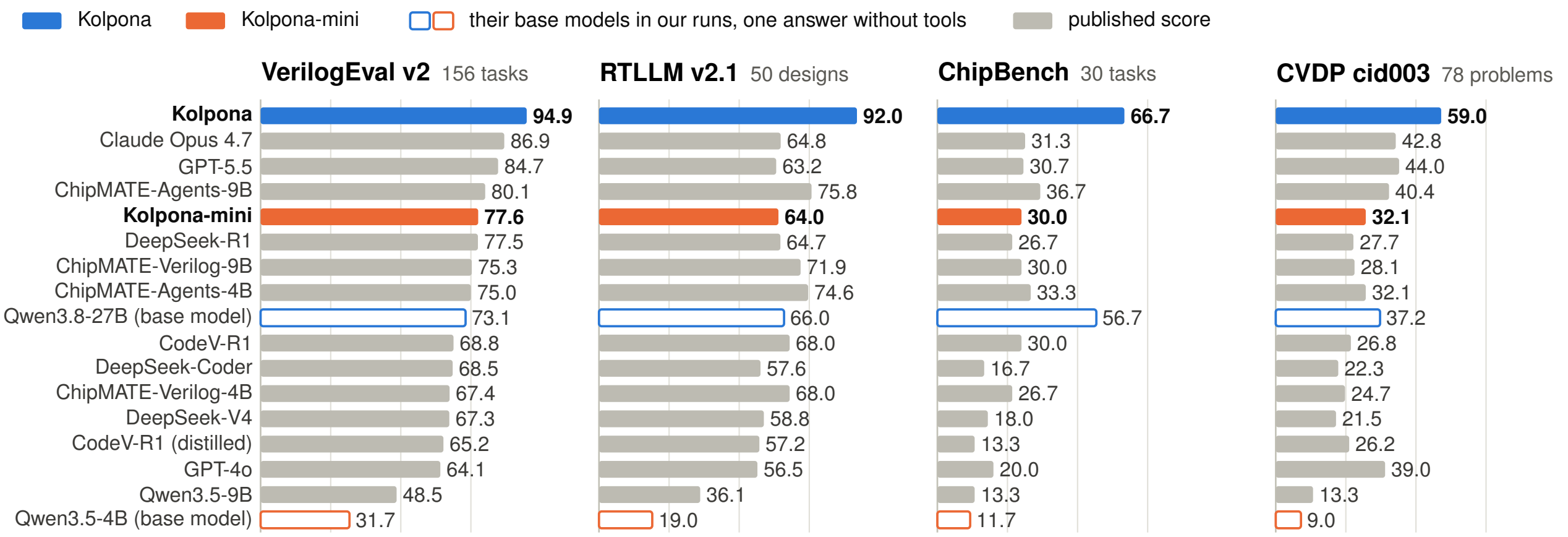

\paragraph{Models.} \sysmini{} is Qwen3.5-4B \citep{qwen2026qwen35_4b} in thinking mode with the edits of \cref{sec:edits}. \sys{} is Qwen3.8-27B \citep{qwen2026qwen38_27b}, run from the checkpoint of \citet{redhat2026qwen38int4}, with 128 spare neurons per MLP and one program. The RTL results of \sys{} come from one NVIDIA RTX 5090 with 32\,GB of memory. Diagnosis and paired tests for the 4B model ran on NVIDIA H100 GPUs and on the same RTX 5090, and editing itself runs on a CPU in minutes. Both sample with temperature 0.6, top-p 0.95 and top-k 20. A base model result is the unedited model with one answer per task and no tools.

\paragraph{Agent budget.} Both systems run $N = 10$ agents per task, except \sysmini{} on VerilogEval v2, which runs $N = 9$, and on the Python benchmarks, which run $N = 5$. Each conversation allows 30 tool calls and 32,768 new tokens per turn, with 96,000 tokens in total for the 27B model and 128,000 for the 4B model.

\paragraph{Practice data.} Diagnosis and paired tests use only practice tasks, namely 2,916 problems of the reinforcement-learning split of CodeV-R1 \citep{zhu2025codevr1} and 10,824 OriGen problems \citep{cui2024origen} whose testbenches were verified against reference designs, both without near-duplicates of VerilogEval tasks.

\paragraph{Benchmarks.} VerilogEval v2 \citep{pinckney2025revisiting} has 156 specification-to-RTL tasks, RTLLM v2.1 \citep{liu2024openllmrtl} has 50 designs and ChipBench \citep{yu2026chipbench} has 30 self-contained (SC) tasks. CVDP \citep{pinckney2025cvdp} contributes the 78 specification-to-RTL problems of category cid003. For Python we use LiveCodeBench v6 \citep{jain2025livecodebench}, HumanEval+ and MBPP+ \citep{chen2021codex,austin2021mbpp,liu2023evalplus}.

\paragraph{Metrics.} Let $\mathcal{T}$ be the tasks of a benchmark, $d_{t,1}$, \dots, $d_{t,N}$ the candidates of task $t$, $\hat{d}_t$ the design that the framework delivers and $v(d) \in \{0, 1\}$ the verdict of the benchmark's testbench on design $d$. We report
\begin{align}
  \text{pass@1} &= \frac{1}{|\mathcal{T}|} \sum_{t \in \mathcal{T}} v(\hat{d}_t), \label{eq:pass1} \\
  \text{single agent} &= \frac{1}{N |\mathcal{T}|} \sum_{t \in \mathcal{T}} \sum_{i=1}^{N} v(d_{t,i}), \label{eq:agent} \\
  \text{best of } N &= \frac{1}{|\mathcal{T}|} \sum_{t \in \mathcal{T}} \max_{i} v(d_{t,i}). \label{eq:best}
\end{align}
Pass@1 scores one run of the full framework. A single agent measures the agents without selection, and best of $N$ bounds any selection from the same candidates. For a base model with $n$ samples per task, pass@1 is the mean of $c_t / n$, where $c_t$ counts the passing samples of task $t$, which is the unbiased estimator of \citet{chen2021codex} at $k = 1$.

\section{Results}

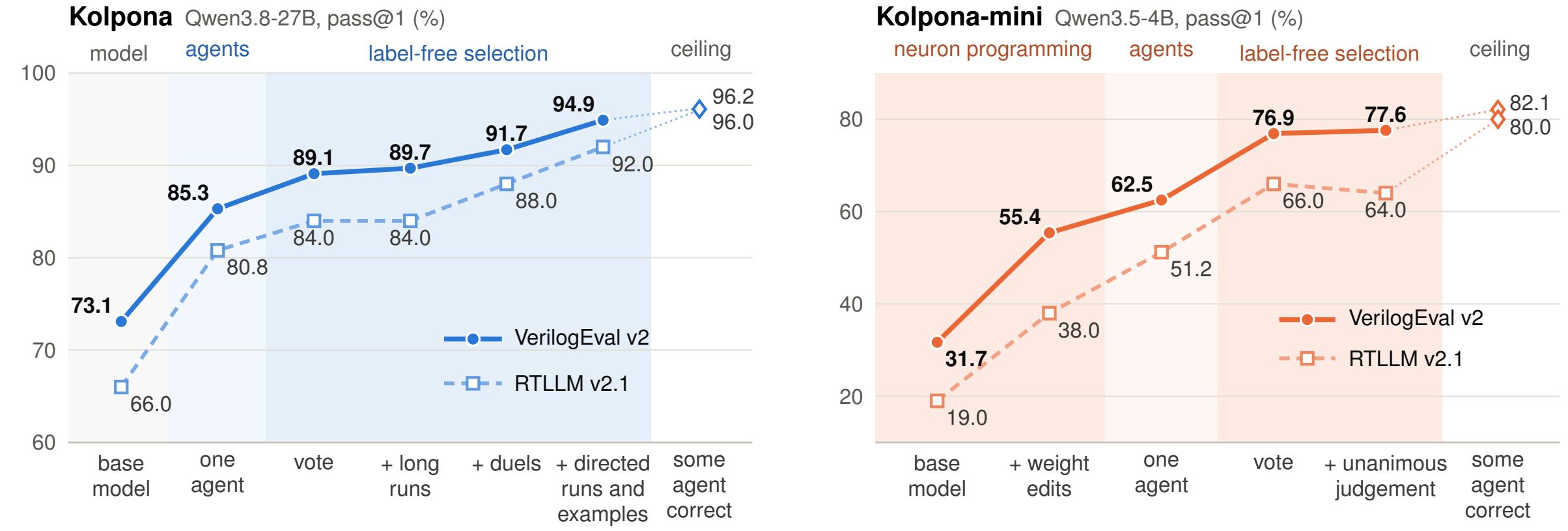
\begin{figure*}[t]
\centering
\begin{tikzpicture}
\pgfplotsset{
  ladder/.style={
    scale only axis, width=6.95cm, height=3.75cm, clip=false,
    axis x line*=bottom, axis y line*=left, y axis line style={draw=none}, x axis line style={draw=kaxis, line width=0.5pt},
    tick style={draw=none}, ymajorgrids, grid style={draw=kgrid, line width=0.4pt},
    yticklabel style={font=\figfontS, text=kmuted}, yticklabel={\pgfmathprintnumber[assume math mode=true]{\tick}},
    xticklabel style={font=\figfontS, text=kinkb, align=center}, set layers,
  },
}
\newcommand{\lval}[4]{\node[font=\figfontS\bfseries, text=kink, anchor=#4, inner sep=2.2pt] at (axis cs:#1,#2) {#3};}
\newcommand{\lvalb}[4]{\node[font=\figfontS, text=kinkb, anchor=#4, inner sep=2.4pt] at (axis cs:#1,#2) {#3};}
\begin{axis}[ladder, name=big, ymin=60, ymax=100, xmin=0.45, xmax=7.55,
    ytick={60,70,80,90,100}, xtick={1,...,7},
    xticklabels={{base\\model},{one\\agent},{vote},{+ long\\runs},{+ duels},{+ directed\\runs and\\examples},{some\\agent\\correct}}]
  \begin{pgfonlayer}{axis background}
  \fill[kink!3] (axis cs:0.45,60) rectangle (axis cs:1.5,100);
  \fill[kblue!6] (axis cs:1.5,60) rectangle (axis cs:2.5,100);
  \fill[kblue!12] (axis cs:2.5,60) rectangle (axis cs:6.5,100);
  \end{pgfonlayer}
  \node[font=\figfontS, text=kmuted, anchor=south] at (axis cs:0.975,100) {model};
  \node[font=\figfontS, text=kblue!80!black, anchor=south] at (axis cs:2,100) {agents};
  \node[font=\figfontS, text=kblue!80!black, anchor=south] at (axis cs:4.5,100) {label-free selection};
  \node[font=\figfontS, text=kmuted, anchor=south] at (axis cs:7.02,100) {ceiling};
  \draw[kblue!55, line width=0.6pt, densely dotted] (axis cs:6,94.9) -- (axis cs:7,96.2);
  \draw[kblue!55, line width=0.6pt, densely dotted] (axis cs:6,92.0) -- (axis cs:7,96.0);
  \addplot[kblue!62, line width=1.1pt, densely dashed, mark=square*, mark size=1.7pt,
           mark options={solid, fill=white, draw=kblue!75, line width=0.8pt}]
    coordinates {(1,66.0) (2,80.8) (3,84.0) (4,84.0) (5,88.0) (6,92.0)};
  \addplot[kblue, line width=1.4pt, mark=*, mark size=1.9pt, mark options={fill=kblue, draw=white, line width=0.6pt}]
    coordinates {(1,73.1) (2,85.3) (3,89.1) (4,89.7) (5,91.7) (6,94.9)};
  \addplot[only marks, mark=diamond*, mark size=2.4pt, mark options={fill=white, draw=kblue, line width=0.8pt}]
    coordinates {(7,96.1)};
  \lval{1}{73.1}{73.1}{south east} \lval{2}{85.3}{85.3}{south east} \lval{3}{89.1}{89.1}{south}
  \lval{4}{89.7}{89.7}{south} \lval{5}{91.7}{91.7}{south} \lval{6}{94.9}{94.9}{south east}
  \lvalb{1}{66.0}{66.0}{north west} \lvalb{2}{80.8}{80.8}{north west} \lvalb{3}{84.0}{84.0}{north}
  \lvalb{4}{84.0}{84.0}{north} \lvalb{5}{88.0}{88.0}{north west} \lvalb{6}{92.0}{92.0}{north west}
  \node[font=\figfontS, text=kinkb, anchor=west, inner sep=3.6pt, align=left] at (axis cs:7,96.1) {96.2\\96.0};
  \draw[kblue, line width=1.4pt] (axis cs:4.35,71.2) -- (axis cs:4.95,71.2);
  \path[fill=kblue, draw=white, line width=0.6pt] (axis cs:4.65,71.2) circle[radius=1.9pt];
  \node[font=\figfontS, text=kink, anchor=west, inner sep=2pt] at (axis cs:5.0,71.2) {VerilogEval v2};
  \draw[kblue!62, line width=1.1pt, densely dashed] (axis cs:4.35,66.4) -- (axis cs:4.95,66.4);
  \path[fill=white, draw=kblue!75, line width=0.8pt] ($(axis cs:4.65,66.4)+(-1.7pt,-1.7pt)$) rectangle ($(axis cs:4.65,66.4)+(1.7pt,1.7pt)$);
  \node[font=\figfontS, text=kink, anchor=west, inner sep=2pt] at (axis cs:5.0,66.4) {RTLLM v2.1};
  \node[anchor=south west, inner sep=0pt] at ($(axis description cs:0,1)+(0cm,0.42cm)$) {\figfontT \sys{}\hspace{1.2mm}{\figfontS\mdseries\color{kmuted}Qwen3.8-27B, pass@1 (\%)}};
\end{axis}
\begin{axis}[ladder, at={(big.east)}, anchor=west, xshift=1.25cm, ymin=10, ymax=90, xmin=0.45, xmax=6.55,
    ytick={20,40,60,80}, xtick={1,...,6},
    xticklabels={{base\\model},{+ weight\\edits},{one\\agent},{vote},{+ unanimous\\judgement},{some\\agent\\correct}}]
  \begin{pgfonlayer}{axis background}
  \fill[korange!14] (axis cs:0.45,10) rectangle (axis cs:2.5,90);
  \fill[korange!6] (axis cs:2.5,10) rectangle (axis cs:3.5,90);
  \fill[korange!14] (axis cs:3.5,10) rectangle (axis cs:5.5,90);
  \end{pgfonlayer}
  \node[font=\figfontS, text=korange!75!black, anchor=south] at (axis cs:1.5,90) {neuron programming};
  \node[font=\figfontS, text=korange!75!black, anchor=south] at (axis cs:3,90) {agents};
  \node[font=\figfontS, text=korange!75!black, anchor=south] at (axis cs:4.5,90) {label-free selection};
  \node[font=\figfontS, text=kmuted, anchor=south] at (axis cs:6.02,90) {ceiling};
  \draw[korange!60, line width=0.6pt, densely dotted] (axis cs:5,77.6) -- (axis cs:6,82.1);
  \draw[korange!60, line width=0.6pt, densely dotted] (axis cs:5,64.0) -- (axis cs:6,80.0);
  \addplot[korange!62, line width=1.1pt, densely dashed, mark=square*, mark size=1.7pt,
           mark options={solid, fill=white, draw=korange!80, line width=0.8pt}]
    coordinates {(1,19.0) (2,38.0) (3,51.2) (4,66.0) (5,64.0)};
  \addplot[korange, line width=1.4pt, mark=*, mark size=1.9pt, mark options={fill=korange, draw=white, line width=0.6pt}]
    coordinates {(1,31.7) (2,55.4) (3,62.5) (4,76.9) (5,77.6)};
  \addplot[only marks, mark=diamond*, mark size=2.4pt, mark options={fill=white, draw=korange, line width=0.8pt}]
    coordinates {(6,82.1) (6,80.0)};
  \lval{1}{31.7}{31.7}{north west} \lval{2}{55.4}{55.4}{south east} \lval{3}{62.5}{62.5}{south east}
  \lval{4}{76.9}{76.9}{south} \lval{5}{77.6}{77.6}{south}
  \lvalb{1}{19.0}{19.0}{north west} \lvalb{2}{38.0}{38.0}{north west} \lvalb{3}{51.2}{51.2}{north west} \lvalb{4}{66.0}{66.0}{north west} \lvalb{5}{64.0}{64.0}{north}
  \node[font=\figfontS, text=kinkb, anchor=west, inner sep=3.2pt] at (axis cs:6,83.6) {82.1};
  \node[font=\figfontS, text=kinkb, anchor=west, inner sep=3.2pt] at (axis cs:6,78.4) {80.0};
  \draw[korange, line width=1.4pt] (axis cs:4.05,36.6) -- (axis cs:4.55,36.6);
  \path[fill=korange, draw=white, line width=0.6pt] (axis cs:4.3,36.6) circle[radius=1.9pt];
  \node[font=\figfontS, text=kink, anchor=west, inner sep=2pt] at (axis cs:4.6,36.6) {VerilogEval v2};
  \draw[korange!62, line width=1.1pt, densely dashed] (axis cs:4.05,28.2) -- (axis cs:4.55,28.2);
  \path[fill=white, draw=korange!80, line width=0.8pt] ($(axis cs:4.3,28.2)+(-1.7pt,-1.7pt)$) rectangle ($(axis cs:4.3,28.2)+(1.7pt,1.7pt)$);
  \node[font=\figfontS, text=kink, anchor=west, inner sep=2pt] at (axis cs:4.6,28.2) {RTLLM v2.1};
  \node[anchor=south west, inner sep=0pt] at ($(axis description cs:0,1)+(0cm,0.42cm)$) {\figfontT \sysmini{}\hspace{1.2mm}{\figfontS\mdseries\color{kmuted}Qwen3.5-4B, pass@1 (\%)}};
\end{axis}
\end{tikzpicture}
\caption{Ablation of the delivered pass@1. Each point adds one stage, and the last point is the share of tasks that some agent solves, an upper bound for any selection. \cref{tab:ablation} lists the numbers.}
\label{fig:ladder}
\end{figure*}

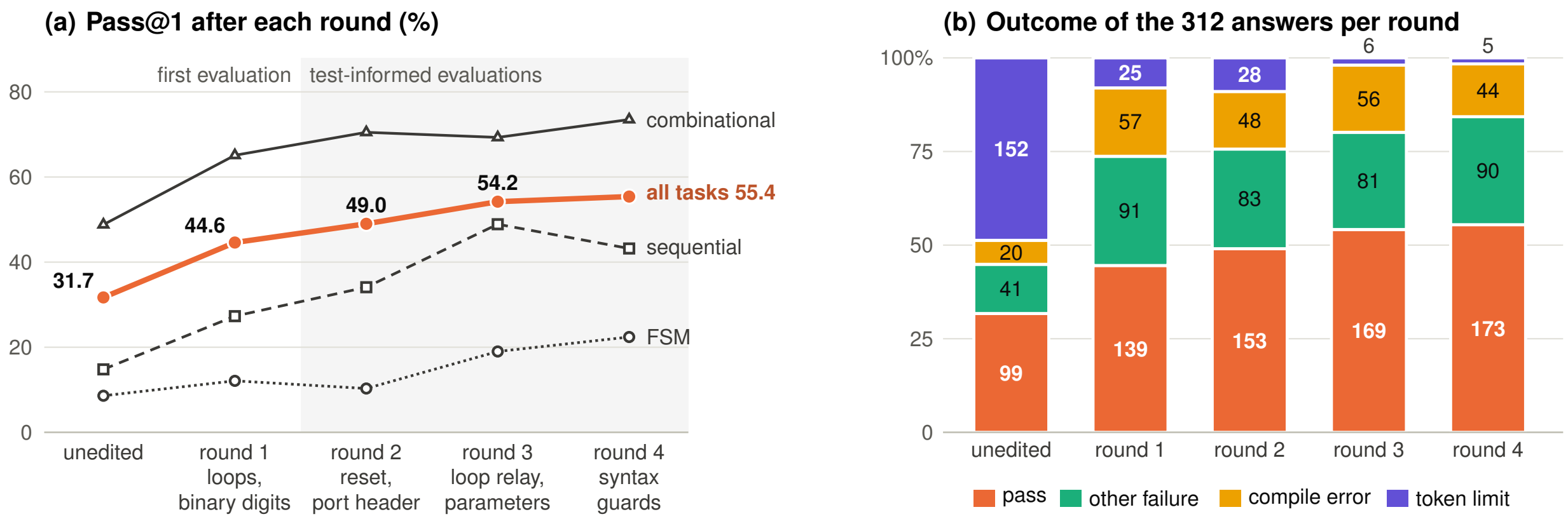
\begin{figure*}[t]
\centering
\begin{tikzpicture}
\begin{axis}[name=pa, scale only axis, width=6.7cm, height=3.9cm, clip=false,
    ymin=0, ymax=88, xmin=-0.45, xmax=4.45, ytick={0,20,40,60,80}, xtick={0,...,4},
    xticklabels={{unedited},{round 1\\loops,\\binary digits},{round 2\\reset,\\port header},{round 3\\loop relay,\\parameters},{round 4\\syntax\\guards}},
    axis x line*=bottom, axis y line*=left, y axis line style={draw=none}, x axis line style={draw=kaxis, line width=0.5pt},
    tick style={draw=none}, ymajorgrids, grid style={draw=kgrid, line width=0.4pt},
    yticklabel style={font=\figfontS, text=kmuted}, yticklabel={\pgfmathprintnumber[assume math mode=true]{\tick}},
    xticklabel style={font=\figfontS, text=kinkb, align=center}, set layers]
  \begin{pgfonlayer}{axis background}
  \fill[kink!4] (axis cs:1.5,0) rectangle (axis cs:4.45,88);
  \end{pgfonlayer}
  \node[font=\figfontS, text=kmuted, anchor=north west, inner sep=2.5pt] at (axis cs:1.5,88) {test-informed evaluations};
  \node[font=\figfontS, text=kmuted, anchor=north east, inner sep=2.5pt] at (axis cs:1.5,88) {first evaluation};
  \addplot[kinkb, line width=0.8pt, mark=triangle*, mark size=1.6pt, mark options={solid, fill=white, draw=kinkb}]
    coordinates {(0,48.8) (1,65.1) (2,70.5) (3,69.3) (4,73.5)};
  \addplot[kinkb, line width=0.8pt, densely dashed, mark=square*, mark size=1.4pt, mark options={solid, fill=white, draw=kinkb}]
    coordinates {(0,14.8) (1,27.3) (2,34.1) (3,48.9) (4,43.2)};
  \addplot[kinkb, line width=0.8pt, densely dotted, mark=*, mark size=1.4pt, mark options={solid, fill=white, draw=kinkb}]
    coordinates {(0,8.6) (1,12.1) (2,10.3) (3,19.0) (4,22.4)};
  \node[font=\figfontS, text=kinkb, anchor=west, inner sep=3pt] at (axis cs:4.05,73.5) {combinational};
  \node[font=\figfontS, text=kinkb, anchor=west, inner sep=3pt] at (axis cs:4.05,43.2) {sequential};
  \node[font=\figfontS, text=kinkb, anchor=west, inner sep=3pt] at (axis cs:4.05,22.4) {FSM};
  \addplot[korange, line width=1.6pt, mark=*, mark size=2.1pt, mark options={fill=korange, draw=white, line width=0.6pt}]
    coordinates {(0,31.7) (1,44.6) (2,49.0) (3,54.2) (4,55.4)};
  \node[font=\figfontS\bfseries, text=kink, anchor=south east, inner sep=2.6pt] at (axis cs:0,31.7) {31.7};
  \node[font=\figfontS\bfseries, text=kink, anchor=south east, inner sep=2.6pt] at (axis cs:1,44.6) {44.6};
  \node[font=\figfontS\bfseries, text=kink, anchor=south, inner sep=3.2pt] at (axis cs:2,49.0) {49.0};
  \node[font=\figfontS\bfseries, text=kink, anchor=south, inner sep=3.2pt] at (axis cs:3,54.2) {54.2};
  \node[font=\figfontS\bfseries, text=korange!80!black, anchor=west, inner sep=3pt] at (axis cs:4.05,56.4) {all tasks 55.4};
  \node[anchor=south west, inner sep=0pt] at ($(axis description cs:0,1)+(0cm,0.22cm)$) {\figfontT (a)\hspace{1mm}Pass@1 after each round (\%)};
\end{axis}
\begin{scope}[shift={($(pa.south east)+(2.75cm,0)$)}, x=1cm, y=0.0125cm]
  \node[anchor=south west, inner sep=0pt] at ($(-0.1,312)+(0cm,0.22cm)$) {\figfontT (b)\hspace{1mm}Outcome of the 312 answers per round};
  \foreach \t in {0,78,156,234,312} { \draw[kgrid, line width=0.4pt] (-0.1,\t) -- (6.35,\t); }
  \foreach \t/\l in {0/0, 78/25, 156/50, 234/75, 312/100\%} { \node[font=\figfontS, text=kmuted, anchor=east, inner sep=1pt] at (-0.16,\t) {\l}; }
  \foreach \s/\p/\o/\c/\l [count=\i from 0] in {unedited/99/41/20/152, round 1/139/91/57/25, round 2/153/83/48/28, round 3/169/81/56/6, round 4/173/90/44/5} {
    \pgfmathsetmacro{\xa}{0.22+\i*1.24}
    \pgfmathsetmacro{\xb}{\xa+0.78}
    \pgfmathsetmacro{\xm}{\xa+0.39}
    \pgfmathsetmacro{\ya}{\p}
    \pgfmathsetmacro{\yb}{\p+\o}
    \pgfmathsetmacro{\yc}{\p+\o+\c}
    \path[fill=korange, draw=white, line width=0.9pt] (\xa,0) rectangle (\xb,\ya);
    \path[fill=kaqua, draw=white, line width=0.9pt] (\xa,\ya) rectangle (\xb,\yb);
    \path[fill=kyellow, draw=white, line width=0.9pt] (\xa,\yb) rectangle (\xb,\yc);
    \path[fill=kviolet, draw=white, line width=0.9pt] (\xa,\yc) rectangle (\xb,312);
    \node[font=\figfontS\bfseries, text=white] at (\xm,\ya/2) {\p};
    \node[font=\figfontS, text=kink] at (\xm,\ya+\o/2) {\o};
    \node[font=\figfontS, text=kink] at (\xm,\yb+\c/2) {\c};
    \ifnum\l>20 \node[font=\figfontS\bfseries, text=white] at (\xm,\yc/2+156) {\l}; \else \node[font=\figfontS, text=kinkb, anchor=south, inner sep=1.2pt] at (\xm,312) {\l}; \fi
    \node[font=\figfontS, text=kinkb, anchor=north, inner sep=2.5pt] at (\xm,0) {\s};
  }
  \draw[kaxis, line width=0.5pt] (-0.1,0) -- (6.35,0);
  \path[fill=korange] (0.22,-62) rectangle (0.44,-48); \node[font=\figfontS, text=kink, anchor=west, inner sep=1.5pt] at (0.46,-55) {pass};
  \path[fill=kaqua] (1.12,-62) rectangle (1.34,-48); \node[font=\figfontS, text=kink, anchor=west, inner sep=1.5pt] at (1.36,-55) {other failure};
  \path[fill=kyellow] (2.78,-62) rectangle (3.00,-48); \node[font=\figfontS, text=kink, anchor=west, inner sep=1.5pt] at (3.02,-55) {compile error};
  \path[fill=kviolet] (4.52,-62) rectangle (4.74,-48); \node[font=\figfontS, text=kink, anchor=west, inner sep=1.5pt] at (4.76,-55) {token limit};
\end{scope}
\end{tikzpicture}
\caption{Qwen3.5-4B alone after each round of weight edits, on VerilogEval v2. (a) Pass@1 overall and by task type. (b) Outcome of the 312 answers per round.}
\label{fig:wsp}
\end{figure*}

\subsection{Comparison with published systems}

\begin{table}[!b]
\caption{Main results in \%. Base is the unedited model without tools. Single agent, pass@1 with its 95\% Wilson interval \citep{wilson1927probable} and best of \(N\) follow \cref{eq:pass1,eq:agent,eq:best}.}
\label{tab:main}
\centering
\footnotesize
\setlength{\tabcolsep}{3.4pt}
\begin{tabular}{@{}lrrrcr@{}}
\toprule
 & Base & Single & & & Best \\
Benchmark & model & agent & Pass@1 & 95\% CI & of \(N\) \\
\midrule
\multicolumn{6}{@{}l}{\textit{\sys{} (Qwen3.8-27B)}} \\
VerilogEval v2 & 73.1 & 85.3 & \textbf{94.9} & 90.2--97.4 & 96.2 \\
RTLLM v2.1 & 66.0 & 80.8 & \textbf{92.0} & 81.2--96.8 & 96.0 \\
ChipBench & 56.7 & 59.7 & \textbf{66.7} & 48.8--80.8 & 83.3 \\
CVDP cid003 & 37.2 & 52.9 & \textbf{59.0} & 47.9--69.2 & 71.8 \\
\midrule
\multicolumn{6}{@{}l}{\textit{\sysmini{} (Qwen3.5-4B)}} \\
VerilogEval v2 & 31.7 & 62.5 & \textbf{77.6} & 70.4--83.4 & 82.1 \\
RTLLM v2.1 & 19.0 & 51.2 & \textbf{64.0} & 50.1--75.9 & 80.0 \\
ChipBench & 11.7 & 28.3 & \textbf{30.0} & 16.7--47.9 & 43.3 \\
CVDP cid003 & 9.0 & 18.6 & \textbf{32.1} & 22.7--43.0 & 41.0 \\
\bottomrule
\end{tabular}
\end{table}

\cref{fig:compare} compares both systems with published results and with their base models, and \cref{tab:main} reports every metric. \sys{} delivers 94.9\% on VerilogEval v2 and 92.0\% on RTLLM v2.1, which is 8.0 and 16.2 points above the best published entries in the comparison. \sysmini{} exceeds ChipMATE-Agents-4B, which applies reinforcement learning to a 4B model, on VerilogEval with 77.6\% against 75.0\%, and it trails on RTLLM with 64.0\% against 74.6\% and on ChipBench with 30.0\% against 33.3\%. On CVDP, \sys{} solves 46 of the 78 problems. Its 59.0\% is 15.0 points above the best published entry and 21.8 points above its base model. \sysmini{} solves 25 of them, which gives 32.1\%, the same score as ChipMATE-Agents-4B. The unedited 27B model already scores 73.1\% and 66.0\% when it answers once without tools, which shows how much of the result rests on the base model. The unedited 4B model reaches only 31.7\% and 19.0\%, mostly because its reasoning runs to the token limit without an answer.

\subsection{Ablation study}

\cref{fig:ladder} decomposes the delivered scores. For \sys{}, agents with tools raise VerilogEval from 73.1\% for the base model to 85.3\% per agent, the vote adds 3.8 points and the later stages add 5.8 points. On RTLLM the selection adds 11.2 points over a single agent. On CVDP the vote reaches 61.5\%, and the later stages cost two problems. The delivered design is correct for 148 of the 150 VerilogEval tasks that some agent solves, and the two misses are misreadings of the specification shared by most agents and by the judging model.

For \sysmini{}, neuron programming and the agents contribute similar amounts on VerilogEval. Neuron programming raises the model alone from 31.7\% to 55.4\%, and the agents with selection raise it to 77.6\%. The vote supplies most of the agent gain. Relative to the vote, the later selection steps of \sysmini{} add one task on VerilogEval and three on CVDP, cost one on RTLLM and leave ChipBench unchanged.

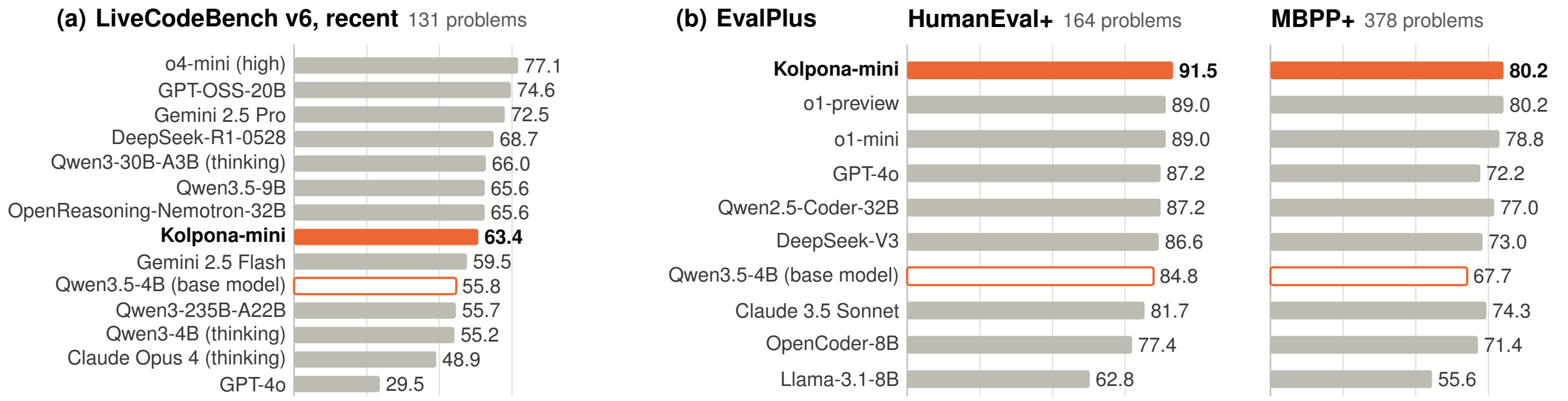
\begin{figure*}[t]
\centering
\begin{tikzpicture}[x=1cm,y=1cm]
  \def\bs{0.032}   %
  \def\na{14}      %
  \def\pa{0.27}    %
  \def\nb{10}      %
  \pgfmathsetmacro{\top}{-0.115}
  \pgfmathsetmacro{\bot}{\top-\na*\pa}
  \pgfmathsetmacro{\pb}{\na*\pa/\nb}
  \node[anchor=south west, inner sep=0pt] at (-2.62,0.12) {\figfontT (a)\hspace{1mm}LiveCodeBench v6, recent\hspace{1mm}{\figfontS\mdseries\color{kmuted}131 problems}};
  \foreach \t in {25,50,75} { \draw[kgrid, line width=0.4pt] (\t*\bs,-0.02) -- (\t*\bs,\bot); }
  \draw[kaxis, line width=0.5pt] (0,-0.02) -- (0,\bot);
  \foreach \n/\v/\k [count=\i] in {o4-mini (high)/77.1/P, GPT-OSS-20B/74.6/P, Gemini 2.5 Pro/72.5/P, DeepSeek-R1-0528/68.7/P,
      Qwen3-30B-A3B (thinking)/66.0/P, Qwen3.5-9B/65.6/P, OpenReasoning-Nemotron-32B/65.6/P, \sysmini{}/63.4/B,
      Gemini 2.5 Flash/59.5/P, Qwen3.5-4B (base model)/55.8/b, Qwen3-235B-A22B/55.7/P, Qwen3-4B (thinking)/55.2/P,
      Claude Opus 4 (thinking)/48.9/P, GPT-4o/29.5/P} {
    \pgfmathsetmacro{\y}{\top-\pa*(\i-0.5)}
    \pgfmathsetmacro{\w}{\v*\bs}
    \node[anchor=east, inner sep=0pt, name\k] at (-0.08,\y) {\n};
    \path[kind\k, rounded corners=0.8pt] (0,\y-0.09) rectangle (\w,\y+0.09);
    \node[anchor=west, inner sep=0pt, val\k] at (\w+0.06,\y) {\v};
  }
  \def\xh{6.75}    %
  \def\xm{10.75}   %
  \node[anchor=south west, inner sep=0pt] at (4.2,0.12) {\figfontT (b)\hspace{1mm}EvalPlus};
  \node[anchor=south west, inner sep=0pt] at (\xh,0.12) {\figfontT HumanEval+\hspace{1mm}{\figfontS\mdseries\color{kmuted}164 problems}};
  \node[anchor=south west, inner sep=0pt] at (\xm,0.12) {\figfontT MBPP+\hspace{1mm}{\figfontS\mdseries\color{kmuted}378 problems}};
  \foreach \x in {\xh,\xm} {
    \foreach \t in {25,50,75} { \draw[kgrid, line width=0.4pt] (\x+\t*\bs,-0.02) -- (\x+\t*\bs,\bot); }
    \draw[kaxis, line width=0.5pt] (\x,-0.02) -- (\x,\bot);
  }
  \foreach \n/\h/\m/\k [count=\i] in {\sysmini{}/91.5/80.2/B,
      o1-preview/89.0/80.2/P, o1-mini/89.0/78.8/P, GPT-4o/87.2/72.2/P, Qwen2.5-Coder-32B/87.2/77.0/P,
      DeepSeek-V3/86.6/73.0/P, Qwen3.5-4B (base model)/84.8/67.7/b, Claude 3.5 Sonnet/81.7/74.3/P, OpenCoder-8B/77.4/71.4/P, Llama-3.1-8B/62.8/55.6/P} {
    \pgfmathsetmacro{\y}{\top-\pb*(\i-0.5)}
    \pgfmathsetmacro{\wh}{\h*\bs}
    \pgfmathsetmacro{\wm}{\m*\bs}
    \node[anchor=east, inner sep=0pt, name\k] at (\xh-0.08,\y) {\n};
    \path[kind\k, rounded corners=0.8pt] (\xh,\y-0.098) rectangle (\xh+\wh,\y+0.098);
    \node[anchor=west, inner sep=0pt, val\k] at (\xh+\wh+0.06,\y) {\h};
    \path[kind\k, rounded corners=0.8pt] (\xm,\y-0.098) rectangle (\xm+\wm,\y+0.098);
    \node[anchor=west, inner sep=0pt, val\k] at (\xm+\wm+0.06,\y) {\m};
  }
\end{tikzpicture}
\caption{General coding, pass@1 (\%). (a) LiveCodeBench v6 problems released from February to April 2025. (b) HumanEval+ and MBPP+ with the extended tests. Published values come from the official leaderboards and model cards.}
\label{fig:general}
\end{figure*}

\cref{fig:wsp} follows the 4B model alone through four rounds of edits, with the one-shot prompt of the benchmark and two samples per task. The first round ends runaway reasoning and raises pass@1 from 31.7\% to 44.6\%. Answers that stop at the token limit fall from 152 to 25 of 312, and the freed answers expose compile errors that later rounds remove. After four rounds the paired gain is 23.7 points, with a 95\% confidence interval from $+17.6$ to $+30.1$ and 63 tasks better against 9 worse, as \cref{tab:wsp} lists. Every edit was built and accepted on practice tasks. The second round targets a reset failure that the first evaluation revealed, so the later evaluations are test-informed. The practice pool confirms the last two rounds independently. They gain 1.3 points on 10,824 paired problems with $p = 0.008$, and 1.5 points on 21,648 paired answers with $z = 4.2$. Finite-state machine (FSM) tasks gain least, because their failures are mostly logic errors that no token-level habit explains.

The edits also help on benchmarks never used while building them. Without tools and with the same seeds, the edited model solves 38.0\% of RTLLM against 19.0\% for the unedited model, a paired gain of 19.0 points with a 95\% confidence interval from $+10.0$ to $+29.0$. On CVDP it solves 17.3\% against 9.0\%, a gain of 8.3 points with an interval from $+2.6$ to $+14.7$, while ChipBench stays at 11.7\%. Its reasoning ends within the token limit in 99\% of the RTLLM answers, against 31\% before the edits.

\subsection{Transfer to general coding}

To see whether the framework is specific to hardware, we also ran it on Python. \sysmini{} runs on Python benchmarks with the same weights and a Python port of the harness, in which the tools run the public examples, edit the code and submit it. Hidden tests are never shown. The agents propose programs, and the same label-free selection as for RTL delivers one of them, with programs grouped by their outputs on shared inputs instead of by simulated behaviour. \cref{fig:general}a covers the 131 LiveCodeBench v6 problems released from February to April 2025, the window that model cards report. \sysmini{} solves 63.4\% of them, against 55.8\% reported for Qwen3.5-4B with one attempt and no tools. On HumanEval+ and MBPP+ it solves 91.5\% and 80.2\% with the extended tests, at or above the best entries of the EvalPlus leaderboard in \cref{fig:general}b, whose models answer once without tools.

The base model of \sysmini{} solves 84.8\% of HumanEval+ and 67.7\% of MBPP+ with one answer and no tools. The edits account for most of this gain, although no Python task was used to build them. With one answer and no tools, the edited model alone solves 91.5\% of HumanEval+ and 77.5\% of MBPP+. It gains 16 problems and loses 5 on HumanEval+, and gains 49 and loses 12 on MBPP+. Exact McNemar tests give $p = 0.027$ and $p < 10^{-5}$. On LiveCodeBench the edited model alone solves 43.5\% against 42.0\% for the unedited model under the same 32,768-token limit, which is within noise, so the gain there comes from the agents.

For \sys{}, HumanEval+ and MBPP+ leave the framework little room. Its base model already solves 93.3\% and 82.8\% with one answer and no tools, above every leaderboard entry, and \sys{} delivers 93.3\% and 82.0\%. \cref{app:general} gives the details. Older problems may be in the training data of every model, and we have no run of the unedited 4B model in the agent harness.

\section{Discussion}

Every edit we kept removes an obstacle that the model already recognizes internally, such as a reasoning loop, a syntax habit or a misread reset requirement. Habits without a linearly readable trigger or a single token at which a program could act, such as timing errors in FSM designs, remained. This agrees with the observation that WSP removes obstacles more readily than it adds abilities \citep{takehi2026wsp}, and \cref{prop:rescue} turns it into a quantitative ceiling.

Selection recovers most of the gap between a single agent and the best of $N$ agents, and its gains depend on the strength of the judging model. Applied to \sysmini{} on VerilogEval, the judgement chain of \sys{} gave no gain, and the 4B judge answered far less reliably. A misreading of the specification that the model and most agents share bounds what any same-model selection can deliver.

\section{Limitations}

Neuron programming needs practice tasks with reference testbenches, so it suits domains in which outputs can be checked by execution. A program can remove only a habit that the model already represents internally and that shows at a single token, and errors in the design logic itself lie beyond such edits. Writing programs still takes diagnosis effort and human supervision. Label-free selection depends on the judging model and on the simulation stimuli. A misreading of the specification that most agents share cannot be recovered, and rare behaviours can escape the stimuli. Running several agents per task costs more inference than a single answer. Public benchmarks may overlap with the pretraining data of the base models, so performance on proprietary designs may differ.

\section{Conclusion}

We presented \sys{}, an agentic framework that improves open-weight models for RTL generation without gradient training. Its neuron programming writes each program into the weights as a single neuron and keeps it only when paired testbench results on fresh practice tasks favour it. On a 4B model, these edits remove recurring failure habits and substantially raise the accuracy of the model on its own. Tool-using agents and label-free selection then deliver one design per task, and with a 27B base model the framework exceeds every published system in our comparison on all four RTL benchmarks while running on a single 32\,GB GPU. The edits also improve Python code generation, although they were built only from RTL practice tasks, which indicates that some of the habits they remove are not specific to hardware. Larger multi-module designs and other domains with executable checks are natural targets for future work.

\bibliography{refs}
\bibliographystyle{icml2026}

\newpage
\appendix
\crefalias{section}{appendix}
\onecolumn

\section{The edits}
\label{app:edits}

\cref{tab:edits} lists the programs of Qwen3.5-4B by role. Locations give the layer and the neuron among the 32 decoder layers. The first three groups act on the model alone, and the last group acts only in agent conversations. \cref{tab:wsp} gives the paired VerilogEval gains of the model alone after each round of edits.

\begin{table}[H]
\caption{Programs of \sysmini{}. A relay passes a marker from an early neuron to a late one. Evidence comes from paired runs on practice tasks unless noted.}
\label{tab:edits}
\centering
\footnotesize
\renewcommand{\arraystretch}{1.16}
\setlength{\tabcolsep}{4pt}
\begin{tabular}{@{}L{2.7cm}L{3.2cm}L{5.8cm}L{4.2cm}@{}}
\toprule
\textbf{Program} & \textbf{Location} & \textbf{Trigger $\to$ effect} & \textbf{Evidence, before $\to$ after} \\
\midrule
\grp{Reasoning control}
Wait suppression & \loc{31}{4356} & always $\to$ \texttt{Wait} $-6$, other hedges held & loop escapes \ba{12}{77} of 170, pass \ba{10.2}{20.3}\% \\
Loop relay & \loc{19}{6263} $\to$ \loc{31}{8873} & detected reasoning loop $\to$ ends the loop & token-limit hits \ba{39}{9} of 512 \\
\grp{Syntax}
Binary digits & \loc{31}{5807} & after a binary base $\to$ digits 2--9 blocked & pass \ba{13.3}{17.2}\% \\
Base flip & \loc{23}{4706} & binary base before a digit 2--9 $\to$ decimal base & 20 of 64 targets flipped \\
Assign guard & \loc{31}{6209} & after \texttt{assign} $\to$ type keywords $-4$ & target class removed \\
Port header & \loc{31}{6905}, \texttt{n8140} & after \texttt{output} in a port list $\to$ \texttt{reg} $+4$, \texttt{logic} $-4$ & duplicate declarations \ba{27}{3} \\
Double typing & \loc{19}{7868} & after \texttt{output reg} $\to$ \texttt{logic} $-16$ & double-typed answers \ba{42}{6} \\
Logic then reg & \loc{31}{8785}, \texttt{n9104} & after \texttt{output logic} $\to$ \texttt{reg} $-2$ & fires on 47 of 96 targets, no ordinary row \\
Parameter relay & \loc{19}{7875} $\to$ \loc{31}{3322} & parameters in a plain port list $\to$ parameter list & pass \ba{39.5}{45.7}\% on 256 tasks \\
Case after assignment & \loc{23}{2597} & after an assignment $\to$ \texttt{case} $-16$ & $P(\texttt{case})$ \ba{0.66}{0.00} \\
Replication & \loc{31}{6408} & concatenation before a replication $\to$ double brace & single-brace replications \ba{230}{76} \\
\grp{Reading the specification}
Reset wording & \loc{19}{1779} & clock-edge decision $\to$ sensitivity list follows the reset wording & asynchronous resets under synchronous wording \ba{52}{21} of 128 \\
\grp{Agent setting}
Turn-start thinking & \loc{31}{7535} & agent turn start $\to$ \texttt{Thinking} $+10$ & thinking opens in every first turn \\
First-turn thinking & \loc{31}{5407}, \texttt{n6941} & first agent turn $\to$ \texttt{Thinking} $+10$ & fires on no later turn start \\
Tool call first & \loc{31}{5636}, \texttt{n7509} & first answer token of the first turn $\to$ \texttt{<tool\_call>} $+12$ & fires on 120 of 120 first turns, no plain prompt \\
Thinking after error & \loc{31}{4430}, \texttt{n3079} & turn start after a compile error $\to$ \texttt{Thinking} $+21.6$ & fires on 113 of 125 such turns, none of 125 others \\
Agent loop escape & \loc{31}{8959}, \texttt{n8786} & loop in an agent turn $\to$ end of thinking $+14$ & loop conversations \ba{23}{6} (VerilogEval run) \\
\bottomrule
\end{tabular}
\end{table}

\begin{table}[H]
\caption{Qwen3.5-4B alone on VerilogEval v2 after each round of edits (one-shot prompt, two samples per task). Gains are paired against the unedited model, with 95\% bootstrap intervals over tasks.}
\label{tab:wsp}
\centering
\footnotesize
\renewcommand{\arraystretch}{1.12}
\setlength{\tabcolsep}{6pt}
\begin{tabular}{@{}llrrlr@{}}
\toprule
Round & Main programs added & Pass@1 & Gain & 95\% CI & Better / worse \\
\midrule
unedited & -- & 31.7 & & & \\
1 & loop control, binary digits & 44.6 & $+12.8$ & [$+7.4$, $+18.6$] & 45 / 12 \\
2 & reset wording, port header & 49.0 & $+17.3$ & [$+11.2$, $+23.4$] & 53 / 10 \\
3 & loop relay, parameter relay & 54.2 & $+22.4$ & [$+16.4$, $+28.5$] & 59 / 9 \\
4 & syntax guards & 55.4 & $+23.7$ & [$+17.6$, $+30.1$] & 63 / 9 \\
\bottomrule
\end{tabular}
\end{table}

The program of \sys{} uses spare neurons 17408 and 17409 of layer 63. It raises \texttt{<tool\_call>} by 12 logits at the first answer token of the first turn, with a linear discriminant trigger computed from first-turn answer starts against later-turn answer starts and positions just before the end of thinking. On 200 practice tasks not used to build it, first answers that open a compile call rose from 181 to 198 and solved tasks stayed at 171, with seven tasks gained and seven lost.

\section{Proofs}
\label{app:proofs}

\paragraph{Proof of \cref{prop:closed}.} For $z \le -\tau \le -1$ we have $\sigma(z) = 1/(1+e^{-z}) \le e^{z}$, so $|\silu(z)| = |z|\,\sigma(z) \le |z|\,e^{z} = f(-z)$ with $f(t) = t\,e^{-t}$. The function $f$ decreases for $t \ge 1$, so $|\silu(z)| \le f(\tau) = \tau e^{-\tau}$. The claim follows from $\|\Delta(x)\| = |\silu(g\T x)|\,|p\T x|\,\|u\|$. For comparison, the minimum of $\silu$ is about $-0.278$ at $z \approx -1.278$. \hfill$\square$

\paragraph{Proof of \cref{prop:rescue}.} The design is extracted from the generated token sequence, and the testbench verdict is a deterministic function of the design. On the complement of $D$ the two sequences coincide, so the designs and the verdicts coincide and $Y_1 = Y_0$. Hence $\E[Y_1 - Y_0] = \E[(Y_1 - Y_0)\mathbf{1}_D] = \Prob(D)\big(\Prob(Y_0=0,Y_1=1\mid D) - \Prob(Y_0=1,Y_1=0\mid D)\big)$. Testbenches with randomized stimuli break the deterministic verdict, and for them the identity holds for the expected verdict. For the bound, the two models compute the same logits at every position before the first opening of a trigger, up to the leak of idle programs. If that leak changes no sampled token, the shared seeds produce the same tokens up to that position, so the sequences can differ only on $F$ and $D \subseteq F$. Since $r + s \le 1$, the identity gives $|\E[Y_1] - \E[Y_0]| \le \Prob(D) \le \Prob(F)$. \hfill$\square$

\paragraph{Proof of \cref{prop:accept}.} With independent tasks, each discordant task favours the edited model with probability $q = r/(r+s)$, so given $m$ discordant tasks $n_{10}$ is $\mathrm{Bin}(m, q)$. If $r \le s$ then $q \le 1/2$. Binomial tails increase in the success probability, so $\Prob_q(n_{10} \ge k) \le \Prob_{1/2}(n_{10} \ge k)$ for every $k$. A test that rejects when the $\mathrm{Bin}(m,1/2)$ tail at the observed $n_{10}$ is at most $\alpha$ therefore rejects with probability at most $\alpha$ for every $m$, and so unconditionally. \hfill$\square$

\section{Harness and selection details}
\label{app:harness}

The agent harness offers three tools, which compile the design, edit it and submit it. The edit tool is introduced by the result of the first compile. The harness has the following features. A final answer that does not compile is replaced by the last design that the compiler accepted. A design that returns with the same tool result four times ends the conversation, which restarts in a fresh context. Guidance cards are appended to a tool result when the result shows their trigger, and every tool result ends with the number of tool calls left. After a successful compile the harness reports ports of the specification that the design lacks, calibrated lint warnings, outputs left unknown on generated inputs and contradictions with plain statements of the specification about the reset and the port widths. The first prompt carries up to three cards of standard design practice chosen by the words of the specification. A reasoning runaway or a context beyond the limit restarts the conversation in a fresh context with a summary of the calls so far, the last design that compiled and the last tool result. The 27B model runs with an 8-bit key-value cache and a context of 100,096 tokens. On VerilogEval, the conversations of \sys{} continued after the first submission with a check against a testbench that the 27B model wrote from the specification, unedited for half of the agents and edited for the other half. Only the first submission enters selection, so this check does not affect any reported result. Agents make few tool calls before they submit. Up to its first submission, an agent of \sys{} makes 2.1 calls on average on VerilogEval, 2.3 on RTLLM and 2.2 on ChipBench, and an agent of \sysmini{} makes 4.1 on RTLLM and 3.7 on ChipBench. The median is two calls in every run, one compile and the submission.

The grouping stimulus is one run of 200 cycles with random inputs in each cycle and a reset about every 16 cycles. The long run has seven phases with about 152,000 cycles in total. They cover fully random inputs, inputs that change with probability 1/8 and 1/128 per cycle, 1-bit inputs held at one and at zero, and windows of constant inputs with a reset pulse. Duels give the model 6,000 thinking tokens for \sys{} and 16,000 for \sysmini{}. \cref{tab:ablation} lists the numbers behind \cref{fig:ladder}.

\begin{table}[H]
\caption{Ablation of the delivered pass@1 (\%), stage by stage. Each column after the single agent adds one stage of selection to the previous one, and best of \(N\) bounds any selection. Base is the unedited model with one answer per task and no tools, and edited is the model of \sysmini{} under the same protocol, on VerilogEval its model-alone state of \cref{tab:wsp}. Judgement is the confirmed knockout for \sys{} and the unanimous rule for \sysmini{}.}
\label{tab:ablation}
\centering
\footnotesize
\setlength{\tabcolsep}{5.2pt}
\begin{tabular}{@{}lrrrrrrr@{}}
\toprule
\multicolumn{8}{@{}l}{\textit{\sys{} (Qwen3.8-27B)}} \\
Benchmark & Base & Single agent & Vote & + long runs & + judgement & + directed runs, examples & Best of \(N\) \\
\midrule
VerilogEval v2 & 73.1 & 85.3 & 89.1 & 89.7 & 91.7 & \textbf{94.9} & 96.2 \\
RTLLM v2.1 & 66.0 & 80.8 & 84.0 & 84.0 & 88.0 & \textbf{92.0} & 96.0 \\
ChipBench & 56.7 & 59.7 & 66.7 & 66.7 & 66.7 & \textbf{66.7} & 83.3 \\
CVDP cid003 & 37.2 & 52.9 & 61.5 & 60.3 & 60.3 & \textbf{59.0} & 71.8 \\
\midrule[\heavyrulewidth]
\multicolumn{8}{@{}l}{\textit{\sysmini{} (Qwen3.5-4B)}} \\
Benchmark & Base & Edited & Single agent & Vote & + long runs & + judgement & Best of \(N\) \\
\midrule
VerilogEval v2 & 31.7 & 55.4 & 62.5 & 76.9 & 76.9 & \textbf{77.6} & 82.1 \\
RTLLM v2.1 & 19.0 & 38.0 & 51.2 & 66.0 & 64.0 & \textbf{64.0} & 80.0 \\
ChipBench & 11.7 & 11.7 & 28.3 & 30.0 & 33.3 & \textbf{30.0} & 43.3 \\
CVDP cid003 & 9.0 & 17.3 & 18.6 & 28.2 & 26.9 & \textbf{32.1} & 41.0 \\
\bottomrule
\end{tabular}
\end{table}

\section{General-coding protocol}
\label{app:general}

The Python harness gives each agent tools that run the public examples, edit the code and submit it, with 32,768 tokens per turn and 128,000 per conversation. $N = 5$ agents with independent seeds propose programs, and the selection of \cref{sec:selection} delivers one of them. \cref{tab:general} lists the results. Leaderboard models on LiveCodeBench were scored on the same 131 problems from the official leaderboard data.

\sys{} runs the same harness with $N = 10$ agents on LiveCodeBench, HumanEval+ and MBPP+. Its tools run the public examples, run a Python program written by the agent and submit, and each conversation may generate 40,000 tokens on LiveCodeBench and 30,000 on the other two. On the 131 LiveCodeBench problems \sys{} delivers 86.3\%, a single agent solves 76.7\% and some agent solves 88.5\%. The same model with empty spare neurons, one answer per problem, no tools and 81,920 tokens solves 78.6\% of these problems. \sys{} solves 14 problems that this run misses and misses 4 that it solves (exact McNemar $p = 0.031$). With 32,768 tokens the model leaves 48 answers without a program and solves 60.3\%. Under that limit and the same settings it solves 93.3\% of HumanEval+ and 82.8\% of MBPP+ with the extended tests, and 97.0\% and 96.3\% with the base tests. Some agent of \sys{} solves 95.7\% and 87.6\% of the problems with the extended tests.

The unedited and the edited Qwen3.5-4B were also run alone, with one answer per problem, no tools, thinking on and 32,768 tokens, on the same problems and with the same seeds. The unedited model solves 84.8\% of HumanEval+, 67.7\% of MBPP+ and 42.0\% of LiveCodeBench, and the edited model solves 91.5\%, 77.5\% and 43.5\%. With the base tests the two models solve 90.2\% and 94.5\% of HumanEval+ and 83.6\% and 92.1\% of MBPP+. On LiveCodeBench 69 answers of the unedited model and 34 of the edited model reach the token limit without a program, which explains why both stay below the 55.8\% of the model card. Every HumanEval+ and MBPP+ program, of any model and system above, passes through the sanitizer of the official EvalPlus code generation script before scoring.

\begin{table}[H]
\caption{\sysmini{} with $N = 5$ agents and \sys{} with $N = 10$ agents on Python benchmarks, pass@1 (\%). HumanEval+ and MBPP+ show base and plus tests, and their reference values are plus-test scores.}
\label{tab:general}
\centering
\footnotesize
\setlength{\tabcolsep}{2.8pt}
\begin{tabular}{@{}lrrrl@{}}
\toprule
Benchmark & Tasks & \sysmini{} & \sys{} & Reference \\
\midrule
LiveCodeBench v6, Feb.--Apr.\ 2025 & 131 & 63.4 & 86.3 & Qwen3.5-4B 55.8, o4-mini (high) 77.1, Gemini 2.5 Flash 59.5 \\
HumanEval+ & 164 & 97.6 / 91.5 & 97.6 / 93.3 & o1-preview 89.0, GPT-4o 87.2 \\
MBPP+ & 378 & 95.0 / 80.2 & 96.8 / 82.0 & o1-preview 80.2, Qwen2.5-Coder-32B 77.0 \\
\bottomrule
\end{tabular}
\end{table}

\section{Worked example}
\label{app:example}

\cref{fig:example} follows \sys{} on VerilogEval task Prob096, a finite-state machine that searches for the sequence 1101. One agent first compiles a design whose five states do not fit its 2-bit state register, so the last state wraps to the first and the machine can never latch. The design compiles, and the automatic check of the harness flags the truncated constant. The agent widens the register and the constants in two edits and submits. Across the $N = 10$ agents, the candidates form two behaviour groups under the vote stimulus. Eight designs stay in the state for the prefix 11 when a third 1 arrives, while two return to the state for 1 and miss the sequence inside 11101. The vote delivers a design of the larger group, no later stage changes the pick, and the delivered design passes the benchmark testbench.

\lstdefinestyle{tool}{basicstyle=\ttfamily\fontsize{6.4}{7.6}\selectfont, columns=fullflexible, keepspaces=true,
  breaklines=true, breakindent=0pt, showstringspaces=false, upquote=true, aboveskip=0pt, belowskip=0pt}

\begin{figure}[H]
\centering
\begin{lrbox}{\exbox}\begin{minipage}{\textwidth}
\begin{tcolorbox}[enhanced, colback=kink!3, colframe=kinkb!70, boxrule=0.5pt, arc=2pt, left=4pt, right=4pt,
  top=1.5pt, bottom=1.5pt, fontupper=\scriptsize, title={Specification \hfill VerilogEval v2, Prob096},
  fonttitle=\scriptsize\bfseries, coltitle=white, colbacktitle=kinkb]
I would like you to implement a module named TopModule with the following interface. All input and output ports are one bit unless otherwise specified. \texttt{input clk}, \texttt{input reset}, \texttt{input data}, \texttt{output start\_shifting}. The module should implement a finite-state machine that searches for the sequence 1101 in an input bit stream. When the sequence is found, it should set start\_shifting to 1, forever, until reset. Reset is active high synchronous. Assume all sequential logic is triggered on the positive edge of the clock.
\end{tcolorbox}
\vspace{1mm}
\begin{minipage}[t]{0.5\textwidth}
\begin{chatbox}{kblue}{Agent\quad\faRobot\ \ compile\_verilog}
\textit{[\ldots] S3 on data==0: last bits ``1100''. [\ldots] No prefix match. S0. Correct. Let me compile to verify.}
\begin{lstlisting}[style=vlog]
localparam S0 = 2'd0,
           S1 = 2'd1,
           S2 = 2'd2,
           S3 = 2'd3,
           S4 = 2'd4;
reg [1:0] state, next_state;   // ... rest of the design
\end{lstlisting}
\end{chatbox}
\begin{chatbox}{kaqua}{Harness}
\begin{lstlisting}[style=tool]
OK
c.sv:12: warning: Numeric constant truncated to 2 bits.
Automatic checks (warnings that often mean wrong behaviour; check each against the specification):
- line 12: vector literal too large for the given number of bits (3 bits needed)
    | S4 = 2'd4;
[introduces the edit tool] (tool calls left: 29)
\end{lstlisting}
\end{chatbox}
\begin{chatbox}{kblue}{Agent\quad\faRobot\ \ edit\_design}
\textit{There's a warning: S4 = 2'd4 doesn't fit in 2 bits (4 needs 3 bits). I was using 5 states with a 2-bit register, but 2 bits can only hold 0--3. I need a 3-bit state register.}
\begin{lstlisting}[style=vlog]
- reg [1:0] state, next_state;
+ reg [2:0] state, next_state;
\end{lstlisting}
\end{chatbox}
\begin{chatbox}{kaqua}{Harness}
\begin{lstlisting}[style=tool]
Edit applied (lines 14-14 of the new design).
Compile result: OK
[the same warning on line 12] (tool calls left: 28)
\end{lstlisting}
\end{chatbox}
\begin{chatbox}{kblue}{Agent\quad\faRobot\ \ edit\_design}
\textit{The localparam constants themselves are 2-bit. Fixing those too.}
\begin{lstlisting}[style=vlog]
- localparam S0 = 2'd0, ..., S4 = 2'd4;
+ localparam S0 = 3'd0, ..., S4 = 3'd4;
\end{lstlisting}
\end{chatbox}
\begin{chatbox}{kaqua}{Harness}
\begin{lstlisting}[style=tool]
Edit applied (lines 8-12 of the new design).
Compile result: OK
\end{lstlisting}
\end{chatbox}
\begin{chatbox}{kblue}{Agent\quad\faRobot\ \ submit}
\textit{Clean compile. The design is complete [\ldots] S4 is a latch state [\ldots] once found, start\_shifting stays 1 until reset.} The submitted design is the agent's candidate.
\end{chatbox}
\end{minipage}\hfill
\begin{minipage}[t]{0.482\textwidth}
\begin{chatbox}{kaqua}{Selection\quad 10 candidates, 2 groups}
\textbf{Group 1}, 8 designs, including this agent's:
\begin{lstlisting}[style=vlog]
// after 111 the input still ends in 11
S2: next_state = data ? S2 : S3;
\end{lstlisting}
\textbf{Group 2}, 2 designs:
\begin{lstlisting}[style=vlog]
// after 111 it falls back to 1
S2: next_state = data ? S1 : S3;
\end{lstlisting}
On the input \texttt{11101}, group 1 raises \texttt{start\_shifting} and group 2 does not. The vote delivers a design of group 1.
\end{chatbox}
\begin{chatbox}{kyellow!80!black}{Delivered design\quad passes the benchmark testbench}
\begin{lstlisting}[style=vlog]
module TopModule
(
  input  clk,
  input  reset,
  input  data,
  output start_shifting
);

  localparam S0 = 3'd0;
  localparam S1 = 3'd1;
  localparam S2 = 3'd2;
  localparam S3 = 3'd3;
  localparam S4 = 3'd4;

  reg [2:0] state;
  reg [2:0] next_state;

  // State register (synchronous, active-high reset)
  always @(posedge clk) begin
    if (reset)
      state <= S0;
    else
      state <= next_state;
  end

  // Next state logic
  always @(*) begin
    case (state)
      S0: next_state = (data == 1'b1) ? S1 : S0;
      S1: next_state = (data == 1'b1) ? S2 : S0;
      S2: next_state = (data == 1'b1) ? S2 : S3;
      S3: next_state = (data == 1'b1) ? S4 : S0;
      S4: next_state = S4;
      default: next_state = S0;
    endcase
  end

  // Moore output
  assign start_shifting = (state == S4);

endmodule
\end{lstlisting}
\end{chatbox}
\end{minipage}
\end{minipage}\end{lrbox}
\scalebox{1.0}{\usebox{\exbox}}
\caption{\sys{} on VerilogEval task Prob096. Left, one agent's conversation, abridged, with the model's own words in italics. Right, the selection among all candidates and the delivered design.}
\label{fig:example}
\end{figure}

\end{document}